\documentclass[a4paper,fleqn]{cas-dc}

\usepackage[authoryear,longnamesfirst]{natbib}
\UseRawInputEncoding

\usepackage{caption}
\usepackage{subcaption}
\usepackage{amsmath,amssymb,amsfonts}
\usepackage{algorithmic}
\usepackage{framed,multirow}
\usepackage{graphicx}
\usepackage{adjustbox}
\usepackage{hyperref}
\usepackage{float}

\usepackage{algorithm2e}
\usepackage{placeins}

\graphicspath{ {./figs/} }

\restylefloat{figure}
\restylefloat{table}

\begin{document}
\let\WriteBookmarks\relax
\def\floatpagepagefraction{1}
\def\textpagefraction{.001}

\shorttitle{CrosScaleNet}


\title{Learning Temporal Saliency for Time Series Forecasting with Cross-Scale Attention}                      

%

\author[1,2]{Ibrahim Delibasoglu}
         [orcid=0000-0001-8119-2873]
         \cormark[1]
\ead{ibrahimdelibasoglu@sakarya.edu.tr}

\author[2]{Fredrik Heintz}[orcid=0000-0002-9595-2471]
\ead{fredrik.heintz@liu.se}






\affiliation[1]{Department of Software Engineering, Faculty of Computer and Information Sciences, Sakarya University,
    city={Sakarya},
    country={Turkiye}}

\affiliation[2]{Department of Computer and Information Science (IDA), REAL, AIICS, Linkoping University,
    city={Linkoping},
    country={Sweden}}

    
\cortext[cor1]{Corresponding author}

\begin{abstract}
Explainability in time series forecasting is essential for improving model transparency and supporting informed decision-making. In this work, we present CrossScaleNet, an innovative architecture that combines a patch-based cross-attention mechanism with multi-scale processing to achieve both high performance and enhanced temporal explainability. By embedding attention mechanisms into the training process, our model provides intrinsic explainability for temporal saliency, making its decision-making process more transparent. Traditional post-hoc methods for temporal saliency detection are computationally expensive, particularly when compared to feature importance detection. While ablation techniques may suffice for datasets with fewer features, identifying temporal saliency poses greater challenges due to its complexity. We validate CrossScaleNet on synthetic datasets with known saliency ground truth and on established public benchmarks, demonstrating the robustness of our method in identifying temporal saliency. Experiments on real-world datasets for forecasting task show that our approach consistently outperforms most transformer-based models, offering better explainability without sacrificing predictive accuracy. Our evaluations demonstrate superior performance in both temporal saliency detection and forecasting accuracy. Moreover, we highlight that existing models claiming explainability often fail to maintain strong performance on standard benchmarks. CrossScaleNet addresses this gap, offering a balanced approach that captures temporal saliency effectively while delivering state-of-the-art forecasting performance across datasets of varying complexity.
\end{abstract}


\maketitle

\section{Introduction}

Time series analysis is a growing area of research with applications in forecasting, classification, imputation, and anomaly detection. Despite advancements, many industry applications require transparent and interpretable models, especially in high-stakes domains. This has driven the demand for Explainable Artificial Intelligence (XAI) in time series forecasting, where understanding the decision-making process is critical. While XAI has seen significant development in image-based tasks, its application to time series data remains less explored. Most XAI efforts in time series have focused on classification, with several reviews discussing interpretability techniques \citep{theissler2022explainable,rojat2021explainable,vsimic2021xai}. However, there is a noticeable gap in XAI for time series forecasting, which requires explaining both feature importance and temporal dependencies.

XAI methods can be categorized as post-hoc vs. ante-hoc, global vs. local, and model-agnostic vs. model-specific. Post-hoc methods, such as SHAP \citep{lundberg2017unified} and LIME \citep{ribeiro2016should}, provide explanations after training and are widely used in regression and classification tasks. However, time series data involves temporal dependencies, requiring explanations that capture sequential relationships. Model-specific methods, such as N-BEATS \citep{oreshkin2019n} and Multilevel Wavelet Decomposition Networks \citep{wang2018multilevel}, embed interpretability directly into the model. Attention mechanisms, as used in Temporal Fusion Transformer (TFT) \citep{lim2021temporal} and Spatio-Temporal Attention LSTM \citep{ding2020interpretable}, have also been explored for interpretability, though their effectiveness is debated \citep{serrano2019attention,jain2019attention}.

Recent architectures in time series forecasting, such as NHITS \citep{challu2023nhits}, iTransformer \citep{liu2023itransformer}, TimeMixer \citep{wang2024timemixer} and LMSAutoTSF \citep{Delibasoglu2024lms}, prioritize accuracy over interpretability. PatchTST \citep{nie2022time} segments time series into patches for multivariate forecasting, while LMSAutoTSF uses multi-scale processing with autocorrelation for temporal processing in the encoders. ETSFormer \citep{woo2202etsformer} is a time-series Transformer architecture that modifies self-attention with exponential smoothing and frequency attention, enabling interpretable decomposition of time-series data into components like level, growth, and seasonality while improving forecasting accuracy and efficiency, making it a suitable choice for comparison in our study. In this paper, we focus on improving explainability in time series forecasting while maintaining competitive accuracy. Fully MLP-based methods like LMSAutoTSF and TimeMixer both achieve state-of-the-art (SOTA) performance by leveraging trend and seasonal decomposition. LMSAutoTSF processes trend and seasonal components across multiple scales, using autocorrelation as a skip connection for temporal processing and combining predictions through simple summation. It also employs decomposition in frequency domain with learnable cutoff frequencies. In contrast, TimeMixer applies a fixed decomposition and uses a mixing strategy, applying a fine-to-coarse approach for seasonal components and a coarse-to-fine approach for trend components to refine details. While TimeMixer introduces a relatively complex refinement strategy, LMSAutoTSF achieves slightly better performance with a simpler approach. However, neither method provides inherent explainability. In contrast, transformer-based methods such as iTransformer and PatchTST exhibit competitive performance on public benchmarks and offer the possibility of explainability through attention map visualization. However, attention mechanisms may focus on inputs due to positional biases or other learned patterns, which might not accurately represent true saliency. Notably, prior works \citep{serrano2019attention,jain2019attention} highlight the limitations of attention mechanisms in providing true explainability. To address this, our work introduces an effective approach aimed at enhancing the saliency-capturing capacity of attention mechanisms through the proposed methodology.

Our primary contribution is addressing this gap by focusing on explainability for time series analysis mostly required for industrial tasks. We compare the interpretability of these models on a synthetic dataset with known feature and time-step importance, and evaluate their accuracy on public benchmarks. Building on the effective trend and seasonal component analysis of LMSAutoTSF and TimeMixer, we adopt a similar multi-scale decomposition approach but aim to provide built-in explainability. Unlike LMSAutoTSF and TimeMixer, which do not incorporate any intrinsic explainability mechanisms, our proposed solution enhances explainability while retaining competitive performance. Our architecture, CrossScaleNet, introduces an advanced patch-based cross-attention mechanism to improve interpretability without sacrificing accuracy. By incorporating cross-attention between scales, we enhance the ability to capture temporal dependencies and provide insights into time-step importance. We evaluate CrossScaleNet alongside LMSAutoTSF, TimeMixer, iTransformer, TFT, and PatchTST, demonstrating that our approach delivers superior explainability and competitive accuracy.

The primary contributions of this work are as follows:
 
\begin{itemize}  
\item Development of a synthetic dataset to evaluate explainability methods in time series forecasting, focusing on temporal and feature importance.  
\item Proposal of CrossScaleNet, an advanced architecture with a patch-based cross-attention mechanism for improved multi-scale processing and interpretability.  
\item The proposed CrossScaleNet effectively captures temporal saliency on synthetic datasets and emerges as the best method, offering explainability and higher accuracy across long-term, and short-term forecasting tasks. 
\end{itemize}

\section{Literature Review}

\textbf{XAI for Image Data.}
Grad-CAM \citep{selvaraju2017grad} explains differentiable classifiers like CNNs by linking predictions to input image regions, making it useful for image classification and object detection. DeepLIFT \citep{shrikumar2017learning} compares neuron activations to a reference, attributing importance to input features, while Integrated Gradients \citep{sundararajan2017axiomatic} integrates gradients along a path between baseline and actual inputs for reliable feature attribution. SmoothGrad \citep{smilkov2017smoothgrad} enhances gradient-based interpretability by averaging gradients over noisy input copies, improving sensitivity maps.

\textbf{XAI for Time Series Classification.}
Time series data introduces temporal dependencies, requiring XAI methods to explain feature importance and their evolution over time. SHAP and LIME have been adapted for time series, while model-specific approaches like multilevel wavelet decomposition networks \citep{wang2018multilevel} and CNN-RNN hybrids \citep{lin2017gcrnn} generate interpretable representations. LEFTIST \citep{guilleme2019agnostic} is a model-agnostic framework for time series classification, and ETSCM \citep{hsu2019multivariate} uses attention to identify important ECG segments. LAXCAT \citep{hsieh2021explainable} employs convolutional attention for multivariate time series classification. Attention mechanisms, popularized by \citep{vaswani2017attention}, are widely used in time series classification \citep{liu2022improving,guo2019exploring,zhou2020domain}. \citep{turbe2023evaluation} proposes a framework with quantitative metrics to evaluate post-hoc interpretability methods for time-series classification.

\textbf{XAI for Time Series Forecasting.}
Time series forecasting has seen fewer XAI developments due to its complexity, requiring models to account for both feature importance and future trajectories. ShapTime \citep{zhang2023shaptime} adapts SHAP for forecasting by incorporating temporal dependencies. Counterfactual explanations \citep{wang2023counterfactual} explore how predictions change with altered inputs, while TFT uses attention to capture temporal dependencies and provide interpretability. \citep{oostvogelinterpretable} explores interpretable deep learning for industrial forecasting, emphasizing trust in black-box models. \citep{troncoso2023new} enhances explainability through visual association rules, and ShapTime \citep{zhang2023shaptime} extends SHAP for temporal dynamics, improving practical applicability. EDformer \citep{chakraborty2024edformer} evaluates post-hoc XAI approaches for time series forecasting using transformer-based methods. While it demonstrates efficiency in interpretability, its overall accuracy on public benchmarks falls short compared to the latest architectures.

\begin{table}[hbt]
\scriptsize
\caption{Characteristics of Synthetic Datasets}
\centering
\begin{tabular}{lccc}
\hline
\textbf{Dataset} & \textbf{\begin{tabular}[c]{@{}c@{}}Important \\ lags\end{tabular}} & \textbf{\begin{tabular}[c]{@{}c@{}}Important \\ Features\end{tabular}} & \textbf{\begin{tabular}[c]{@{}c@{}}Target Noise \\ level\end{tabular}} \\
\hline
SYN1 & 1-15 & \{0, 1\} & 0.01 \\
\hline
SYN2 & \begin{tabular}[c]{@{}c@{}}1-5, 9-10, 15-16, 18, 20,\\ 25-26, 35-36, 50-52, 91-95\end{tabular} & \{0, 2\} & 0.05 \\
\hline
SYN3 & \begin{tabular}[c]{@{}c@{}}9-10, 15-16, 18, 20-25,\\ 31, 34, 60-65\end{tabular} & \{1, 2\} & 0.08 \\
\hline
SYN4 & \begin{tabular}[c]{@{}c@{}}9-10, 15-16, 18-21,\\ 41-42, 45-46\end{tabular} & \{1, 2\} & 0.10 \\
\hline
SYN5 & 71-77 & \{1, 2\} & 0.06 \\
\hline
SYN6 & 48-57 & \{0, 2\} & 0.05 \\
\hline
SYN7 & 60, 62-69 & \{0, 1\} & 0.02 \\
\hline
SYN8\textsuperscript{*} & mixture of SYN5 and SYN6 & \{0, 1, 2\} & 0.11 \\
\hline
\end{tabular}
\label{table:dataset_details}
\end{table}

\section{Synthetic Time Series Generation}  
\label{section:syhthetic_dataset}  

To evaluate the performance of time series models and saliency explanations, eight synthetic datasets with different characteristics are generated using pre-defined temporal dependencies, feature importance, and corresponding saliency maps. Consider a dataset \( \mathbf{X} \in \mathbb{R}^{n_{\text{samples}} \times n_{\text{features}}} \), where each feature \( X_{i} \) is categorized as either important or non-important. The target variable \( y \) is constructed using both the current values of important features and their lagged values.  

Each dataset contains a unique combination of \textbf{Important Features} (indices of features with direct and lagged contributions to \( y \)), \textbf{Important Lags} (the set of lag values influencing the target variable), and \textbf{Noise Level} (standard deviation of Gaussian noise added to \( y \)). The saliency maps of datasets named \textit{SYN1, SYN2, SYN3, SYN4, SYN5, SYN6, SYN7}, and \textit{SYN8} exhibit distinct temporal focusing patterns, reflecting variability in the temporal importance across datasets as shared in Table \ref{table:dataset_details}. The first four datasets give more importance to recent historical points, while the others give more importance to the oldest historical points. This variability is crucial for evaluating the temporal saliency capability of models, particularly in assessing how well a model can identify and focus on the most relevant time steps for predictions.

\begin{figure*}[!ht]
\centering
  \includegraphics[width=0.8\textwidth]{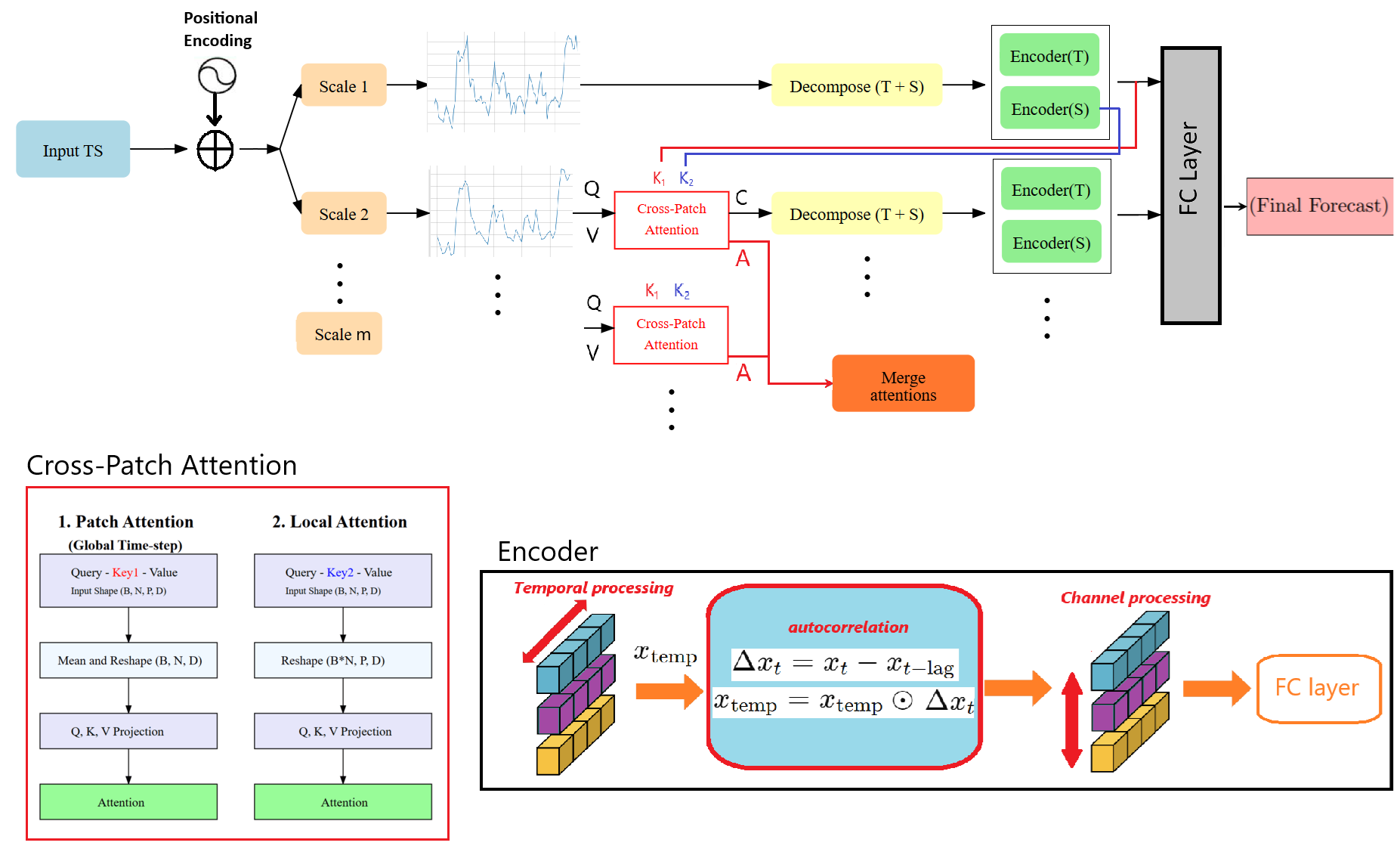}
  \caption{General overview of the multi-scale processing with patch-based cross-attention}
\label{fig:overview}
\end{figure*}

\section{Methodology}
\label{section:methodology}

Given a $D$-dimensional multivariate time series $\mathbf{X} = [\mathbf{x}_1, \mathbf{x}_2, \dots, \mathbf{x}_T] \in \mathbb{R}^{D \times T}$ of sequence length $T$, where $\mathbf{x}_t \in \mathbb{R}^D$ represents the data at the $t$-th time step and $D$ denotes the number of features, we aim to develop an explainable time series forecasting model. Our methodology offers intrinsic explainability by leveraging learned attention weights, enabling multi-scale processing that effectively captures and interprets the temporal importance of different time steps.

We introduce a patch-based cross-attention mechanism to capture distinct temporal dependencies, as shown in Figure~\ref{fig:overview}. To ensure interaction across scales, cross-attention is applied between the highest-level merged and the seasonal component outputs. Our architecture combines multi-scale processing with efficient attention to capture temporal patterns at various granularities. The input series is decomposed into trend and seasonal components, processed independently across scales using dedicated encoders. The outputs are merged for the final forecast, with attention maps interpolated and aggregated to provide a comprehensive view of temporal importance.

\subsection{Multi-Scale Processing}

The proposed model processes the input \( X \) at multiple scales \( m \in \{1, \ldots, M\} \). Scales except the highest one are refined using proposed cross-patch attention. Input data are decomposed into seasonal and trend components in each scale:

\begin{equation}
X^m_{\text{seasonal}}, X^m_{\text{trend}} = \text{Decomp}(X^m).
\end{equation}

The decomposed components are processed through their respective Encoders including Fully Connected ($FC$) layers for both temporal and channel processing in the Encoder as detailed in Figure \ref{fig:overview}  to compute scale output $y^m$:

\begin{equation}
\text{y}^m = \text{Encoder}_{\text{Seasonal}}(X^m_{\text{seasonal}}) + \text{Encoder}_{\text{Trend}}(X^m_{\text{trend}}).
\end{equation}

Each output is scaled using a learnable scale weight \( \sigma^m \):

\begin{equation}
\text{y}^m = \text{y}^m \cdot \sigma^m, \quad \text{where} \; \sigma^m = \text{sigmoid}(\text{w}^m).
\end{equation}

Finally, the predictions across all scales are combined to produce the final output $y$ via last $FC$ layer:

\begin{equation}
\text{y} = FC ( \left[  \text{y}^1 , \text{y}^2, .. \text{y}^m   \right]
\end{equation}

\subsubsection{Cross-attention via scales}

For each scale, the model computes predictions and combines them to generate the final forecast via a $FC$ layer. Notably, attention is only applied for scales \( m > 1 \), where the predictions from the first scale (\( \mathbf{y}_1 \)) are used as the keys for the attention module. For all subsequent scales (\( m > 1 \)), the cross-patch attention mechanism refines the input window \( X^m \) by incorporating the predictions from the first scale. This prediction is appropriately scaled and interpolated to match the temporal resolution of \( X^m \), updating the context ($C$) in alignment with the method described in section \ref{section:cross-patch}.

The attention $A$ for scale \( m > 1 \) is computed as follows:

\begin{equation}
A^m = \text{CrossPatchAttention}(Q^m, K1, K2, V^m),
\end{equation}

where:
\begin{itemize}
    \item \( Q^m = V^m = X^m \): The Query and Value matrices derived from the input at scale \( m \).
    \item \( K1 = \text{y}^1 \): The key matrix derived by interpolating the first scale's prediction $y^1$.
    \item \( K2 = (\text{y}_{seasonal}^1\): The key matrix derived by interpolating the first scale's seasonal encoder prediction $y_{season}^1$.
\end{itemize}

\begin{figure}[hbt]
\centering
  \includegraphics[width=0.5\textwidth]{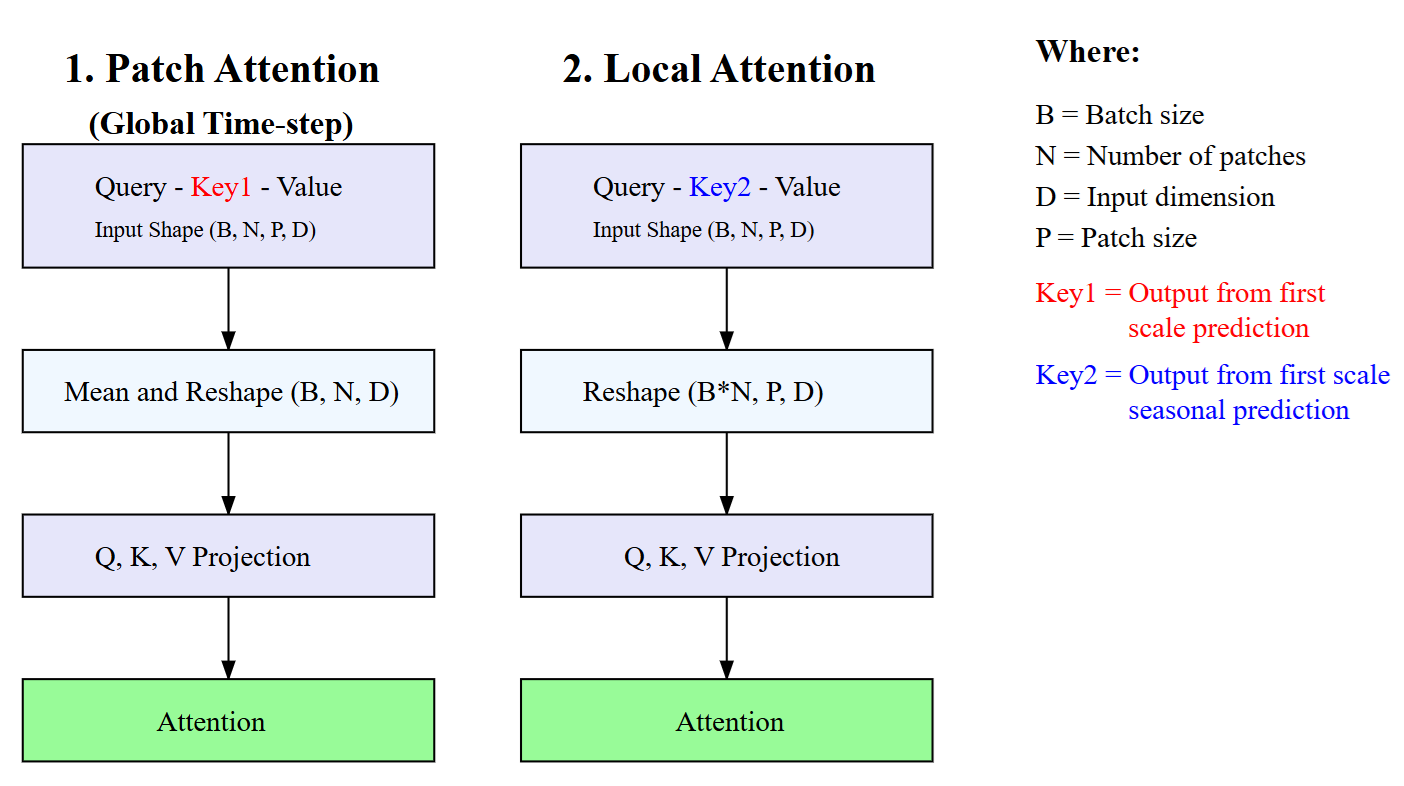}
  \caption {1. Patch Attention processes global temporal relationships by reshaping input into (B, N, D) format, enabling attention computation across patches, 2. Local Attention captures fine-grained patterns by operating within each patch.}
 \vspace{1.5em}
\label{fig:patch_attention}
\end{figure}

\begin{figure}[hbt]
\centering
  \includegraphics[width=0.45\textwidth]{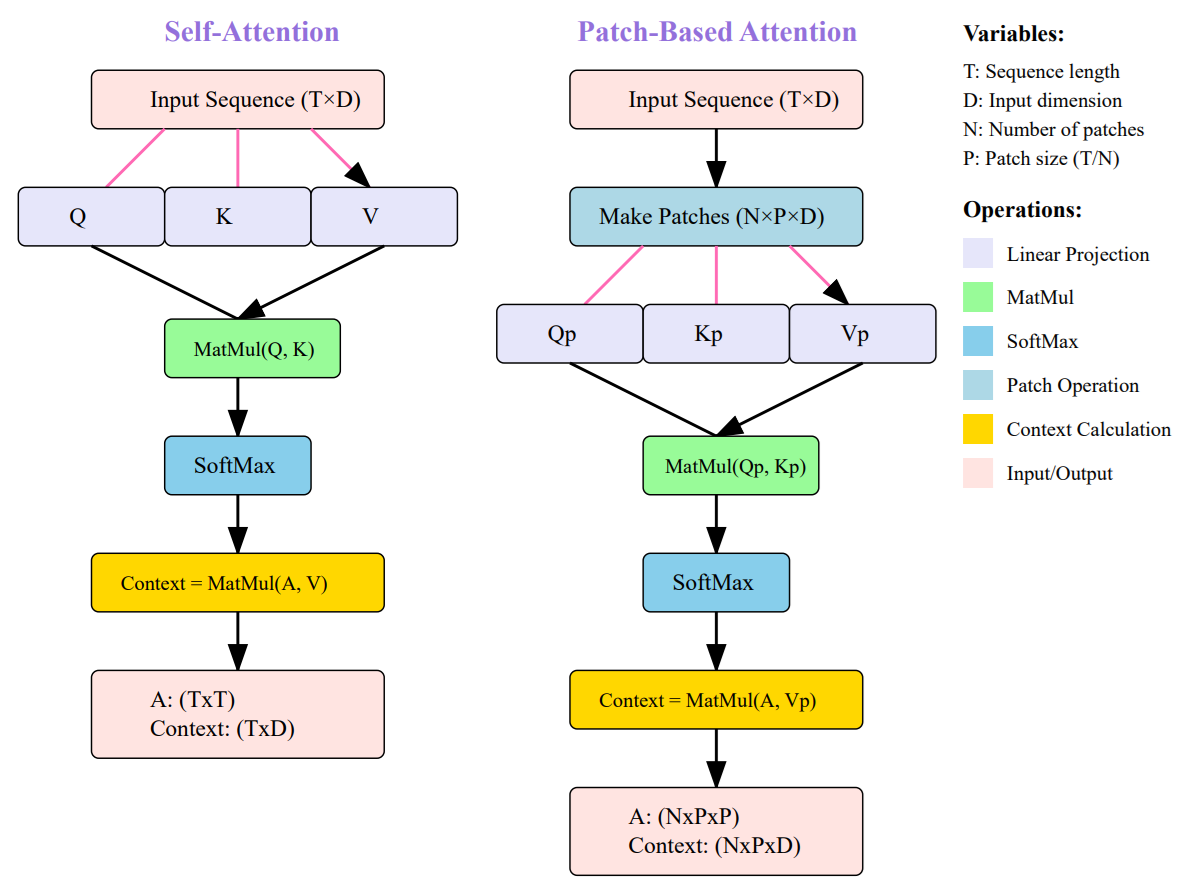}
  \caption{Comparison between Self-Attention and Patch-Based Attention mechanisms. Self-Attention (left) processes the entire sequence directly, projecting it into Query (Q), Key (K), and Value (V) representations. Patch-Based Attention (right) first segments the input into fixed-size patches before computing attention, enabling more efficient processing of long sequences.}
\vspace{1.5em}
\label{fig:self_and_patch_attention}
\end{figure}

The attention mechanism is implemented using patch-based approach, which enhances computational efficiency and better captures temporal patterns by processing patches of the input sequence. Figure \ref{fig:self_and_patch_attention} illustrates the key difference between regular self-attention and the patch-based attention approach. Figure \ref{fig:patch_attention} illustrates the proposed Cross-Patch Attention mechanism, which integrates two complementary attention components:
\begin{itemize}  
    \item \textbf{Patch Attention:} Processes global temporal relationships by reshaping the input into the format (B, N, D), where B is the batch size, N is the number of patches, and D is the feature dimension. This enables attention computation across patches, capturing long-range dependencies in the data.

    \item \textbf{Local Attention:} Focuses on fine-grained patterns within each patch by reshaping the input to (B*N, P, D), where P is the patch length. This allows the model to capture detailed, localized information within individual patches.
\end{itemize}  
The red and blue colored variables in Figure \ref{fig:patch_attention}: \textcolor{red}{Key1} ($K1$) and \textcolor{blue}{Key2} ($K2$), represent distinct inputs for the key component of the attention mechanism. These inputs play a critical role in computing attention scores, ensuring that both global and local patterns are effectively integrated. Together, these components enable the model to achieve a balance between capturing broad temporal trends and fine-grained details, enhancing its performance in time-series tasks.

\subsection{Cross-Patch Attention Mechanism}
\label{section:cross-patch}

Cross-patch attention mechanism is designed to handle hierarchical information. The base attention computation is defined as:

\begin{equation}
\text{Attention}(Q, K, V) = \text{Softmax}\left(\frac{Q K^\top}{\sqrt{d}}\right) V
\end{equation}

where \( Q, K, V \) are the query, key, and value representations, and \( d \) is a scaling factor to normalize attention scores. The mechanism employs two distinct attention modules: \textit{Patch Attention} and \textit{Local Attention}, each explained below.

\subsubsection{Patch Attention}  
The \textit{Patch Attention} mechanism captures relationships between patches across the entire sequence to model global dependencies. The input sequence is divided into patches:

\begin{equation}
X \rightarrow X_p \in \mathbb{R}^{B \times N \times P \times D}
\end{equation}

where \( B \) is the batch size, \( N \) is the number of patches, \( P \) is the patch size, and \( D \) is the input dimension. 

To compute patch-level attention, mean pooling is applied over the patch dimension to extract patch-level representations for the query (\( \text{Pr}_q \)), key (\( \text{Pr}_k \)), and value (\( \text{Pr}_v \)):

\begin{equation}
\text{Pr}_q = \frac{1}{P} \sum_{i=1}^{P} X_{p,q}^{(i)}, \quad 
\text{Pr}_k = \frac{1}{P} \sum_{i=1}^{P} X_{p,k}^{(i)}, \quad 
\text{Pr}_v = \frac{1}{P} \sum_{i=1}^{P} X_{p,v}^{(i)}
\end{equation}

The \textbf{key} representation for Patch Attention (\( X_{p,k} \)) is derived from the \textit{prediction output from the highest-scale} $y^1$, representing the prediction from seasonal and trend encoders for highest scale. This design choice ensures that the global dependencies across patches are captured using the most comprehensive temporal representation. The attention computation proceeds as follows:

\begin{equation}
Q_p = W_q \, \text{Pr}_q, \quad 
K_p = W_k \, \text{Pr}_k, \quad 
V_p = W_v \, \text{Pr}_v,
\end{equation}

where \( W_q, W_k, W_v \in \mathbb{R}^{D \times D} \) are learnable weight matrices. The patch-level attention ($A_P$) is computed as:

\begin{equation}
A_P = Softmax\left(\frac{Q_p K_p^\top}{\sqrt{D}}\right) V_p
\end{equation}

The output of Patch Attention represents the aggregated global context information across patches.

\subsubsection{Local Attention}  
The \textit{Local Attention} mechanism focuses on capturing fine-grained relationships within each patch. Each patch representation (\( X_p \)) is reshaped for local attention computation as:

\begin{equation}
X_L \in \mathbb{R}^{(B \times N) \times P \times D}
\end{equation}

The \textbf{key} representation for Local Attention (\( X_{l,k} \)) is derived from the \textit{prediction output from the highest-scale seasonal component} $y_{season}^1$, representing the prediction from seasonal encoder for highest scale which captures the periodic patterns within the sequence. By using this specialized representation, the Local Attention mechanism emphasizes the relationships between time steps that contribute to seasonal variations. The attention computation is defined as:

\begin{equation}
Q_L = W_{lq} X_L, \quad K_L = W_{lk} X_L, \quad V_L = W_{lv} X_L
\end{equation}

\begin{equation}
A_L = Softmax\left(\frac{Q_L K_L^\top}{\sqrt{D}}\right) V_L
\end{equation}

The result of Local Attention captures the within-patch context, allowing the model to resolve temporal relationships specific to short-term dependencies.

\subsubsection{Cross-Patch Attention Context Fusion}  

The outputs of the Patch Attention and Local Attention mechanisms are combined to generate the final representation. The patch-level context from the Patch Attention mechanism is computed as:

\begin{equation}
C_P = A_P \cdot V_p, \quad \text{where} \quad C_P \in \mathbb{R}^{B \times N \times D}
\end{equation}

To align with the temporal resolution of the sequence, \( C_P \) is reshaped and broadcasted across the time steps within each patch:

\begin{equation}
 C_P  \rightarrow C_P \in \mathbb{R}^{B \times N \times P \times D}
\end{equation}

Similarly, the local context from the Local Attention mechanism is defined as:

\begin{equation}
C_L = A_L \cdot V_L, \quad \text{where} \quad C_L \in \mathbb{R}^{B \times N \times P \times D}
\end{equation}

The final output of the Cross-Patch Attention output \textbf{(C)} is obtained by combining the broadcasted patch context and the local context:

\begin{equation}
\text{C} =  C_P + C_L
\end{equation}
\subsection{Theoretical Motivation for Cross-Scale Attention and Interpretability}

The cross-scale attention mechanism in our model enhances interpretability by aligning hierarchical temporal abstractions with attention-driven relevance signals. Unlike conventional self-attention which compares positions within the same input scale, our approach introduces a semantic alignment across scales by comparing temporal patches at lower scales with abstracted predictions from higher scales. This design enables the model to focus on relevant patterns across resolutions, grounded in learned predictive structure rather than raw input signals.

\paragraph{Interpretability via Hierarchical Temporal Abstraction.}
Each scale \( m \) represents a different level of temporal granularity. Lower scales preserve fine-grained, high-frequency patterns, while higher scales encode coarse, long-range dynamics. Interpretability improves when the model can connect these low-level signals to high-level abstractions. Formally, we define a hierarchy of abstraction functions \( h_m: \mathbb{R}^{D \times T} \rightarrow \mathbb{R}^{D \times T_m} \), where \( T_m < T_{m-1} \), and each abstraction is associated with a corresponding prediction \( y^m \). These hierarchical outputs serve as interpretable anchors for multi-resolution attention.

\paragraph{Cross-Patch Attention as Semantic Matching.}
Our patch-based cross-attention compares localized temporal patches at scale \( m \) against semantic keys derived from predictions \( y^1 \) (and its seasonal component \( y^1_{\text{seasonal}} \)), rather than using the input itself as in self-attention. This key innovation grounds the attention mechanism in semantically meaningful patterns—e.g., trends or periodicities—rather than arbitrary correlations in the input. As a result, attention scores reflect meaningful temporal relevance, improving transparency of the model's decision process. By comparing each patch to higher-level predictive contexts, the model learns to associate specific temporal regions with their abstract outcomes.

\paragraph{Information Bottleneck Interpretation.}
From an information-theoretic perspective, this cross-scale conditioning acts as a form of relevance filtering via a soft information bottleneck. Let \( I(X^m; y^m) \) denote the mutual information between the input at scale \( m \) and its forecast. Conditioning this mapping on the higher-level output \( y^1 \), which serves as a compact, semantically-rich representation, yields a constrained dependency:

\[
I(X^m; y^m \mid y^1) < I(X^m; y^m)
\]

This inequality reflects a desirable regularization: only the components of \( X^m \) that align with the abstract, generalizable structure captured in \( y^1 \) contribute significantly to prediction. In practice, this filtering mechanism suppresses noisy or redundant attention patterns and sharpens focus on temporally salient regions, which leads to more interpretable and stable importance scores across scales.

\begin{table*}[t]
\centering
\scriptsize
\captionsetup{justification=centering}
\caption{Detailed dataset descriptions for forecasting datasets}
\begin{tabular}{|c|c|c|c|c|}
\hline
Category    & Dataset      & Dim & Frequency & Information         \\ \hline
Long-term   & ETTh1, ETTh2 & 7   & Hourly    & Electricity         \\
            & ETTm1, ETTm2 & 7   & 15 min    & Electricity         \\
            & Weather      & 21  & 10 min    & Weather             \\
            & Electricity  & 321 & Hourly    & Electricity         \\
            & Traffic      & 862 & Hourly    & Transportation      \\
            & Exchange     & 8   & Daily     & Economy             \\ 
            & Solar-Energy     & 137   & 10 min     & Electricity   \\
            & Beijing PM2.5     & 12   & Hourly     & Meteorological   \\ \hline
Short-term  & PEMS03       & 358 &  5 min     & Transportation \\
            & PEMS04       & 307 &  5 min     & Transportation \\
            & PEMS07       & 883 &  5 min     & Transportation \\
            & PEMS08       & 170 &  5 min     & Transportation \\ \hline
\end{tabular}
\label{table-dataset}
\end{table*}

\section{Experimental results}

We assess temporal saliency on synthetic datasets with known temporal importance, comparing CrossScaleNet to iTransformer, PatchTST and TFT. While TFT emphasizes interpretability and PatchTST uses patch-based processing, CrossScaleNet distinguishes itself through its multi-scale cross-attention mechanism. We visualize attention weights of PatchTST and iTransformer to analyze their temporal processing behavior. We also assess feature importance on the synthetic dataset using feature ablation and integrated gradients, and evaluate temporal saliency through comprehensiveness and sufficiency metrics. We evaluate forecasting accuracy on real-world benchmark datasets for both short-term PEMS dataset \citep{chen2001freeway} and long-term datasets used by Autoformer \citep{wu2021autoformer}. The Beijing PM2.5 Dataset \citep{beijing_pm2.5_381} features multiple inputs and a single target, unlike others. We use it to analyze temporal saliency in multiple-input, single-target predictions.

\subsection{Dataset Descriptions}

Table \ref{table-dataset} represents the details of datasets used in this study. The PEMS dataset \citep{chen2001freeway}, which is used for short-term traffic flow forecasting, includes four public traffic network datasets: PEMS03, PEMS04, PEMS07, and PEMS08. It provides an additional challenge with its highly dynamic and rapidly changing time series. We also evaluate the performance of our proposed architecture on eight widely-used public benchmark datasets, including Weather, Electricity, Traffic, Exchange and four ETT datasets (ETTh1, ETTh2, ETTm1, ETTm2) for long-term forecasting used by Autoformer \citep{wu2021autoformer}. The Electricity Transformer Temperature (ETT) dataset contains 7 factors of electricity transformer measurements spanning with four subsets: ETTh1 and ETTh2 recorded hourly, and ETTm1 and ETTm2 recorded every 15 minutes. The Exchange dataset comprises daily exchange rates from 8 countries between 1990 and 2016. Additional datasets include Weather (21 meteorological factors collected every 10 minutes), Electricity Consumption Load (ECL) with hourly data from 321 clients, and Traffic data consisting of hourly road occupancy rates from 862 sensors in the San Francisco Bay area.

\subsection{Ablation Studies}

\begin{figure*}[bt]
  \centering
  \begin{subfigure}{0.45\textwidth}  
    \centering
    \includegraphics[width=\textwidth]{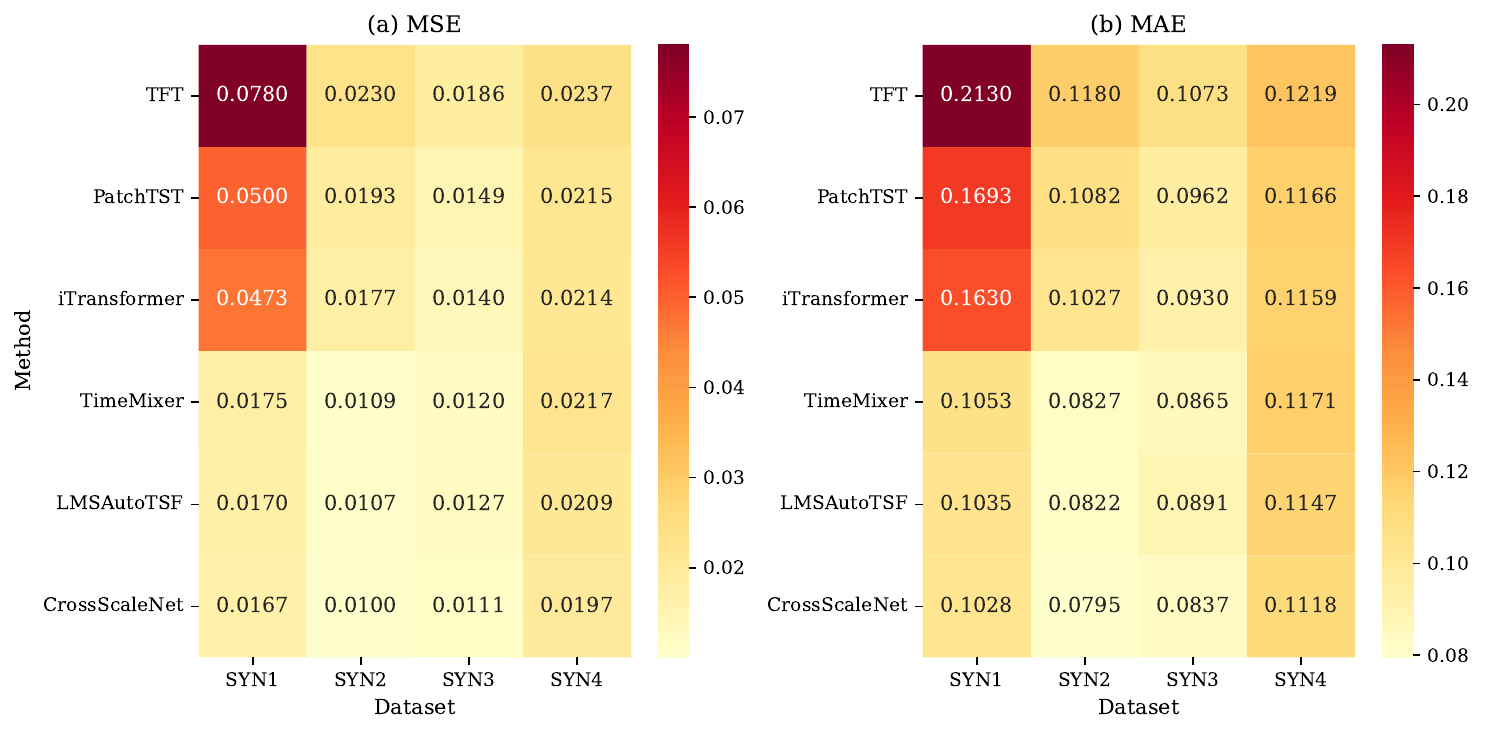}
  \end{subfigure}
  \hspace{0.05\textwidth}  
  \begin{subfigure}{0.45\textwidth}  
    \centering
    \includegraphics[width=\textwidth]{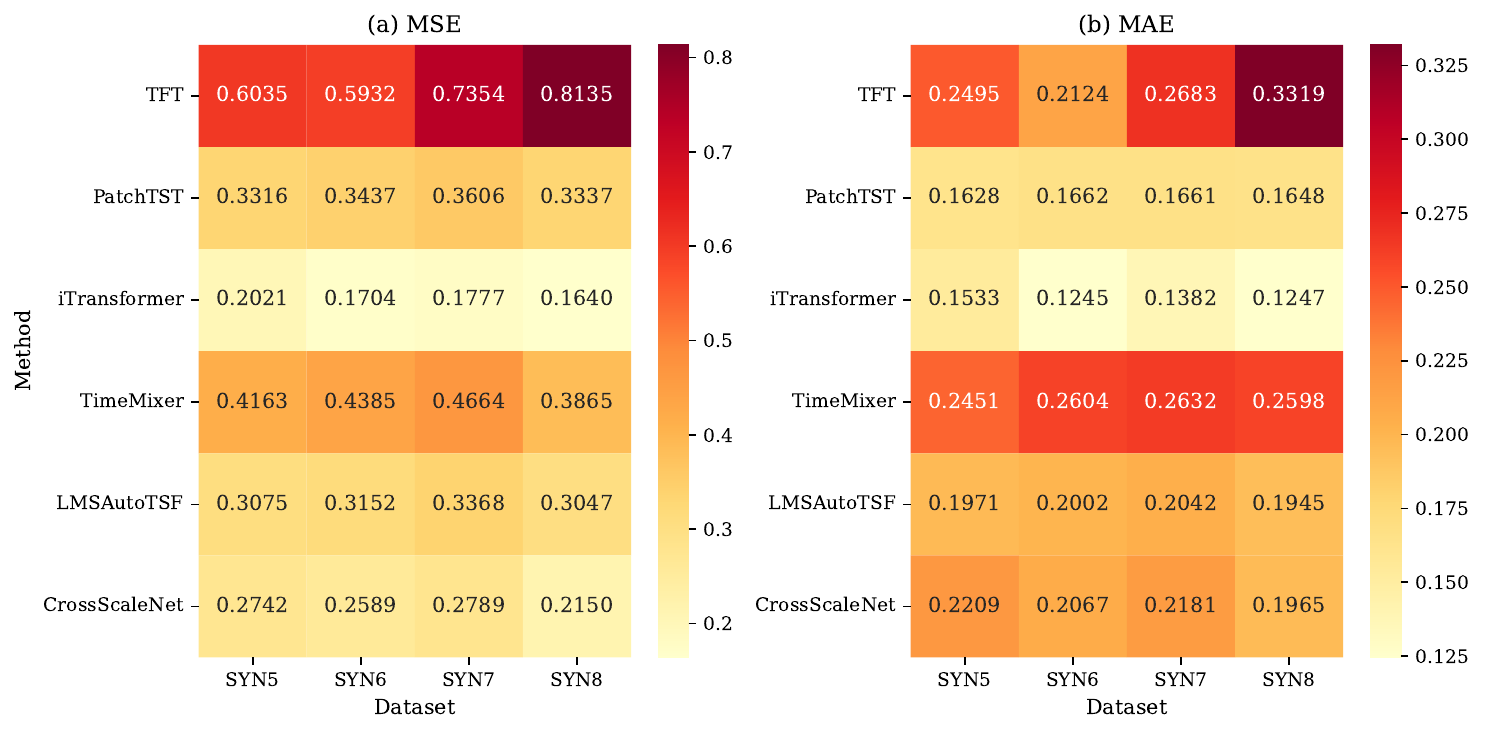}
  \end{subfigure}
  \caption{Performance comparison for forecasting accuracy on synthetic datasets.}
  \label{fig:synth_comparison}
\end{figure*}

Table \ref{table:ablation} presents an ablation study comparing the proposed architecture's performance with two baseline attention mechanisms: self-attention and patch-based attention. In this study, Patch-cross$^{key}$ refers to the proposed cross-attention mechanism, where the same key is shared for both patch and local attention. Meanwhile, Patch-cross$^{key}*$ represents the variant where different keys are used for patch and local attention. Although Patch-cross$^{key}*$ yields relatively lower performance for some synthetic datasets, our experiment focuses on temporal saliency. The results demonstrate that using the seasonal component output as a key for the local patch is more effective in capturing temporal saliency. The results clearly demonstrate the effectiveness of the proposed Cross-Patch Attention mechanism, which combines multi-scale processing and cross-attention to enhance the modeling of temporal dependencies. Consistent improvements in performance are observed across multiple synthetic datasets (SYN1 - SYN8), as indicated by the evaluation metrics: Mean Squared Error (MSE) and Mean Absolute Error (MAE). Performance metrics (MSE, MAE) for the synthetic dataset are shown in Figure \ref{fig:synth_comparison}, alongside comparisons to TFT, PatchTST, iTransformer, TimeMixer, and LMSAutoTSF.

\begin{table}[htbp]
\centering
\caption{Ablation study for CrossScaleNet}
\label{table:ablation}
\begin{tabular}{llcc}
\hline
Dataset    & Model          & MSE       & MAE       \\ \hline
\multirow{3}{*}{SYN1} & Self-Attention& 0.3022     & 0.4686     \\
& Patch-attention    & 0.0187     & 0.1077     \\
& Patch-cross$^{key}$ & 0.0167     & 0.1027     \\
& Patch-cross$^{key}*$ & 0.0167     & 0.1026     \\
&    & \multicolumn{1}{l}{} & \multicolumn{1}{l}{} \\ \hline
SYN2       & Self-Attention & 0.1290     & 0.3005     \\
& Patch-attention    & 0.0152     & 0.0964     \\
& Patch-cross$^{key}$ & 0.0099     & 0.0792     \\
& Patch-cross$^{key}*$ & 0.0099     & 0.0793     \\
&    & \multicolumn{1}{l}{} & \multicolumn{1}{l}{} \\ \hline
SYN3       & Self-Attention & 0.0137     & 0.0921     \\
& Patch-attention    & 0.0135     & 0.0916     \\
& Patch-cross$^{key}$ & 0.112     & 0.0837     \\
& Patch-cross$^{key}*$ & 0.111     & 0.0836     \\
&    & \multicolumn{1}{l}{} & \multicolumn{1}{l}{} \\ \hline
SYN4       & Self-Attention & 0.0196     & 0.1114     \\
& Patch-attention    & 0.0315     & 0.1399     \\
& Patch-cross$^{key}$ & 0.0197     & 0.1118     \\
& Patch-cross$^{key}*$ & 0.0197    & 0.1118     \\
&    & \multicolumn{1}{l}{} & \multicolumn{1}{l}{} \\ \hline
SYN5       & Self-Attention & 0.3469     & 0.2441     \\
& Patch-attention    & 0.2809     & 0.2314     \\
& Patch-cross$^{key}$ & 0.2572     & 0.2194     \\
& Patch-cross$^{key}*$ & 0.2742     & 0.2209     \\
&    & \multicolumn{1}{l}{} & \multicolumn{1}{l}{} \\ \hline
SYN6       & Self-Attention & 0.3451    & 0.2495     \\
& Patch-attention    & 0.3048     & 0.2363     \\
& Patch-cross$^{key}$ & 0.2429     & 0.2043     \\
& Patch-cross$^{key}*$ & 0.2589     & 0.2067     \\
&    & \multicolumn{1}{l}{} & \multicolumn{1}{l}{} \\ \hline
SYN7       & Self-Attention & 0.3116     & 0.2357    \\
& Patch-attention    & 0.3227     & 0.2513     \\
& Patch-cross$^{key}$ & 0.2618    & 0.2165     \\
& Patch-cross$^{key}*$ & 0.2788     & 0.2180     \\
&    & \multicolumn{1}{l}{} & \multicolumn{1}{l}{} \\ \hline
SYN8       & Self-Attention & 0.3308     & 0.2374     \\
& Patch-attention    & 0.2612     & 0.2246     \\
& Patch-cross$^{key}$ & 0.1899     & 0.1888     \\
& Patch-cross$^{key}*$ & 0.2149     & 0.1964     \\
&    & \multicolumn{1}{l}{} & \multicolumn{1}{l}{} \\ \hline
\end{tabular}
\end{table}

\begin{table*}[hbt]
\centering
\caption{Comparison of Sufficiency and Comprehensiveness across Datasets}
\begin{tabular}{|c|c|ccc|ccc|}
\hline
Dataset & Ratio & \multicolumn{3}{c|}{Sufficiency (Lower is better)} & \multicolumn{3}{c|}{Comprehensiveness (Higher is better)} \\
\cline{3-8}
 & & CrossScaleNet & iTransformer & PatchTST & CrossScaleNet & iTransformer & PatchTST \\ \hline

\multirow{3}{*}{Electricity} & 10\% & \textcolor{red}{0.591} & 0.724 & 0.677 & \textcolor{red}{0.167} & 0.132 & 0.101 \\
 & 20\% & \textcolor{red}{0.522} & 0.634 & 0.611 & \textcolor{red}{0.242} & 0.227 & 0.175 \\
 & 50\% & \textcolor{red}{0.331} & 0.393 & 0.371 & 0.430 & \textcolor{red}{0.496} & 0.418 \\ \hline

\multirow{3}{*}{PM2.5} & 10\% & \textcolor{red}{0.484} & 0.674 & 0.55 & \textcolor{red}{0.373} & 0.076 & 0.139 \\
 & 20\% & \textcolor{red}{0.449} & 0.664 & 0.537 & \textcolor{red}{0.366} & 0.104 & 0.157 \\
 & 50\% & 0.389 & 0.656 & \textcolor{red}{0.203} & 0.432 & 0.175 & \textcolor{red}{0.519} \\ \hline
\end{tabular}
\label{tab-suff_vs_comp}
\end{table*}

\begin{table*}[hbt]
\centering
\caption{Comparison of models on PM2.5 Dataset with multivariate input single target. (prediction horizons (12,24,48) and fixed look-back 24)}
\label{tab:pm25_comparison}
\centering
\begin{tabular}{|c|cc|cc|cc|cc|cc|}
\hline
  & \multicolumn{2}{c}{{CrossScaleNet}} & \multicolumn{2}{c}{{LMSAutoTSF}} & \multicolumn{2}{c}{TimeMixer} & \multicolumn{2}{c}{iTransformer} & \multicolumn{2}{c}{PatchTST}\\ \hline
 & MSE  & MAE  & MSE  & MAE   & MSE  & MAE  & MSE  & MAE  & MSE   & MAE   \\
\hline
12  & \textcolor{blue}{0.3722}  & \textcolor{blue}{0.3789}  & \textcolor{red}{0.3654}    & \textcolor{red}{0.3783}   & 0.3731   & 0.3791  & 0.4108 & 0.4024 & 0.4202  & 0.4093  \\
24  & \textcolor{blue}{0.6059}  & \textcolor{blue}{0.5068}  & \textcolor{red}{0.5936}    & \textcolor{red}{0.5053}   & 0.6402   & 0.5254  & 0.6510 & 0.5294 & 0.6586  & 0.5275  \\
48  & \textcolor{red}{0.9088}  & \textcolor{red}{0.6518}  & \textcolor{blue}{0.9125}    & \textcolor{blue}{0.6557}   & 0.9271   & 0.6623  & 0.9767 & 0.6751 & 0.9753  & 0.6728  \\
\hline
\textbf{Avg.} & \textcolor{blue}{0.6289} & \textcolor{red}{0.5125} & \textcolor{red}{0.6238} & \textcolor{blue}{0.5131} & 0.6468 & 0.5223 & 0.6795 & 0.5356 & 0.6847 & 0.5365 \\ \hline
\end{tabular}
\end{table*}

\begin{figure*}[ht] 
  \centering
  \begin{subfigure}{.30\textwidth}
    \centering 
    \includegraphics[width=\textwidth]{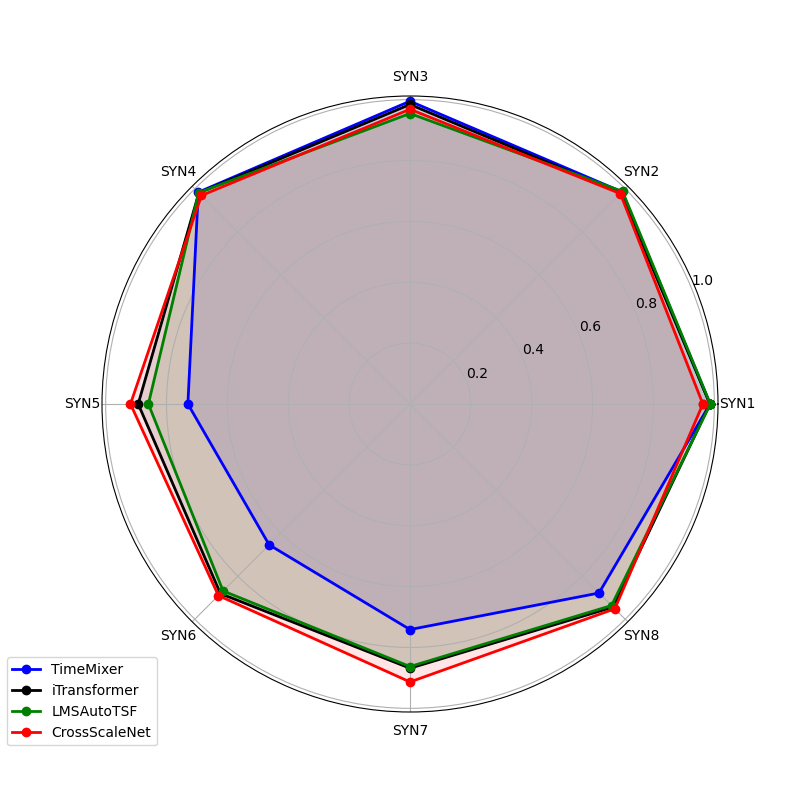}
    \caption{Feature ablation}
  \end{subfigure}
  \hspace{0.05\textwidth} 
  \begin{subfigure}{.30\textwidth}
    \centering
    \includegraphics[width=\textwidth]{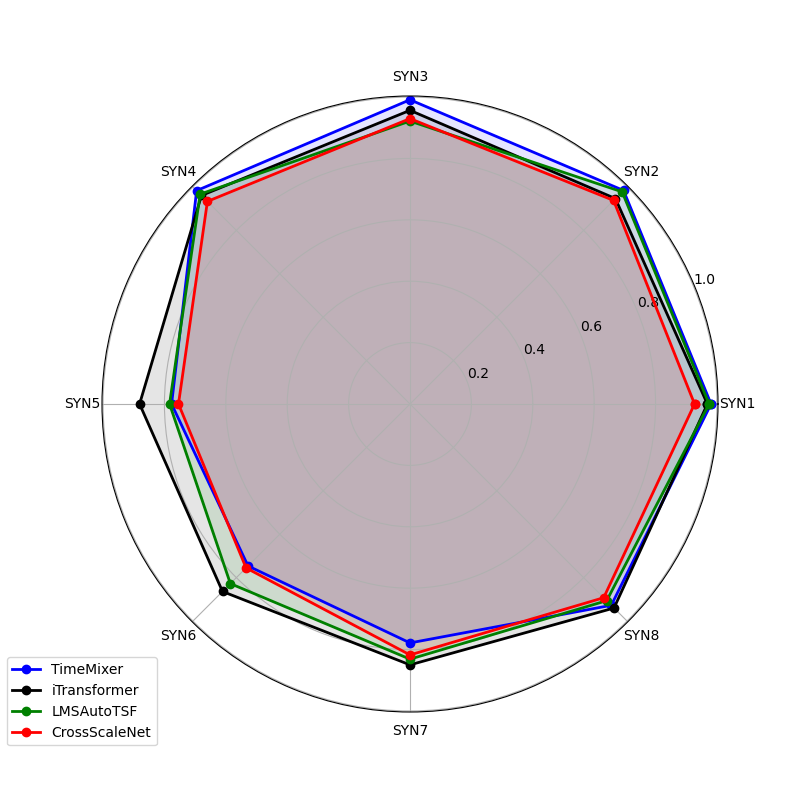}
    \caption{Integrated Gradients}
  \end{subfigure}
  \vspace{1.5em}
  \caption{Comparison of important feature scores for each dataset with different methods via feature ablation.}
  \label{fig:feature_imp_simple}
\end{figure*}

\begin{figure}[hbt]
    \centering
        \begin{subfigure}[b]{0.2\textwidth}
            \centering
            \includegraphics[width=\textwidth]{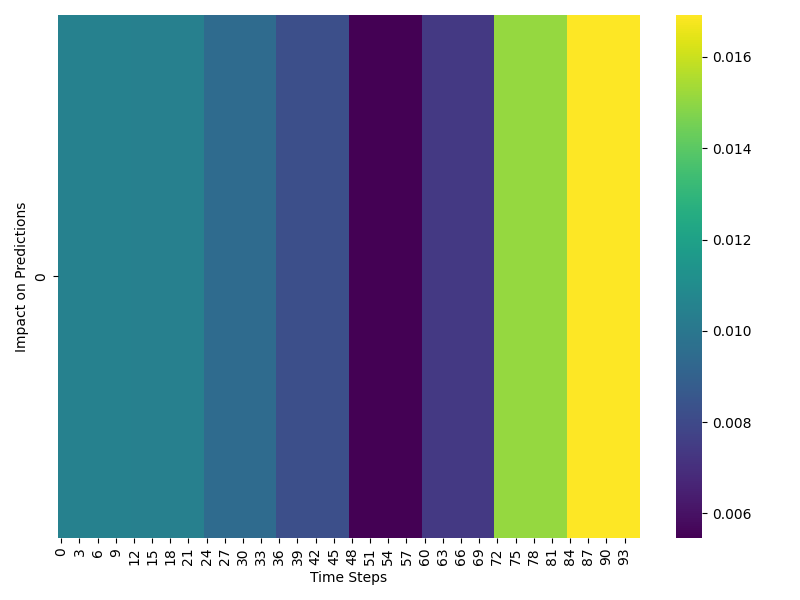}
            \caption{Shaptime}
            \label{fig:1a}
        \end{subfigure}
        \hfill
        \begin{subfigure}[b]{0.2\textwidth}
            \centering
            \includegraphics[width=\textwidth]{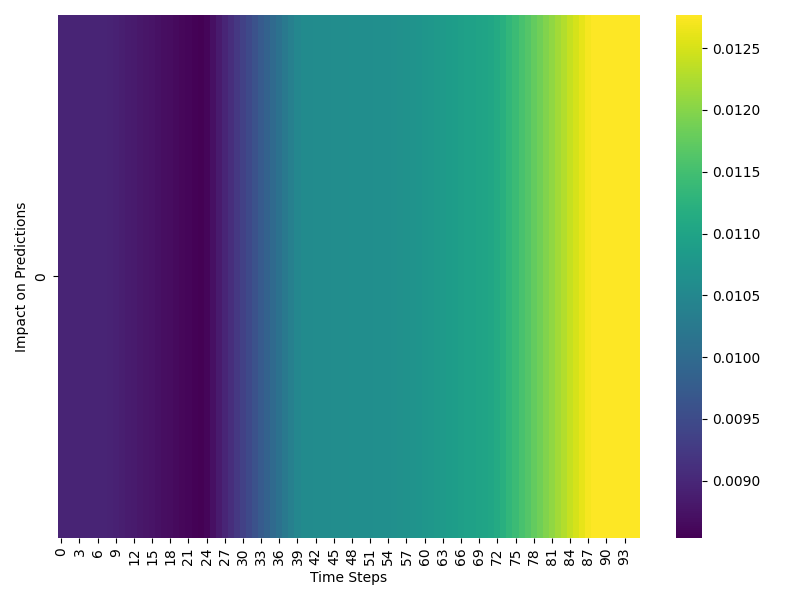}
            \caption{CrossScaleNet (ours)}
            \label{fig:1b}
        \end{subfigure}
        \vskip\baselineskip 
        \begin{subfigure}[b]{0.2\textwidth}
            \centering
            \includegraphics[width=\textwidth]{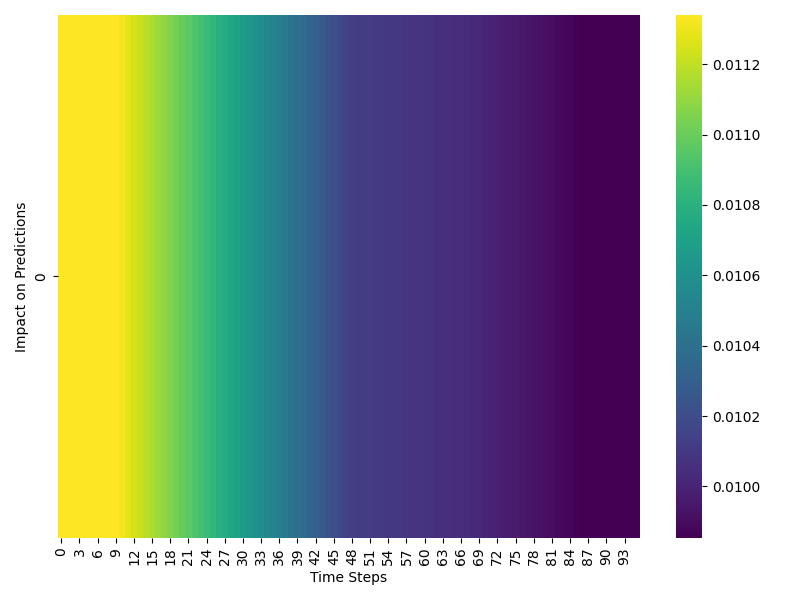}
            \caption{iTransformer}
            \label{fig:1c}
        \end{subfigure}
        \hfill
        \begin{subfigure}[b]{0.2\textwidth}
            \centering
            \includegraphics[width=\textwidth]{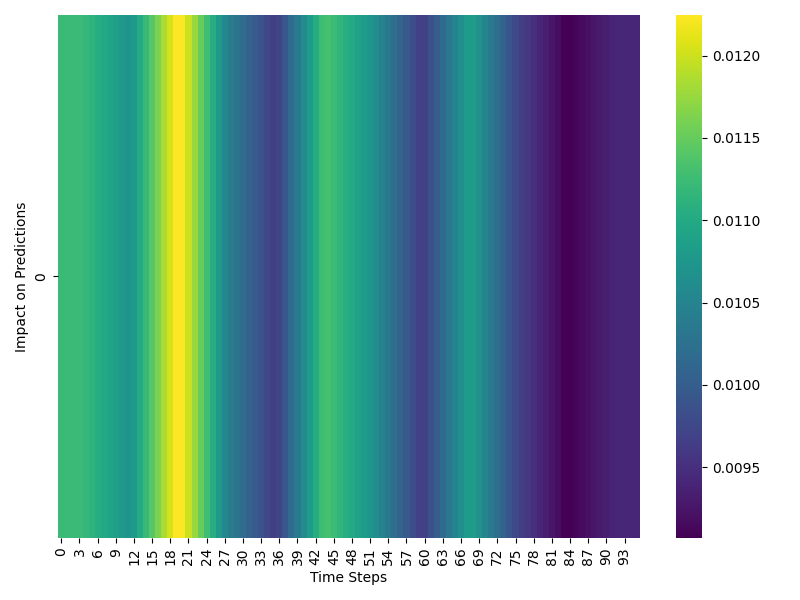} 
            \caption{PatchTST}
            \label{fig:1d}
        \end{subfigure}
\vspace{1.5em}
\caption{Temporal saliency for ETTh1 sample prediction}
\label{fig:sample_attentions}
\end{figure}

\begin{figure*}[htbp]
    \centering
    \begin{subfigure}[t]{0.25\textwidth}
        \centering
        \includegraphics[width=\textwidth]{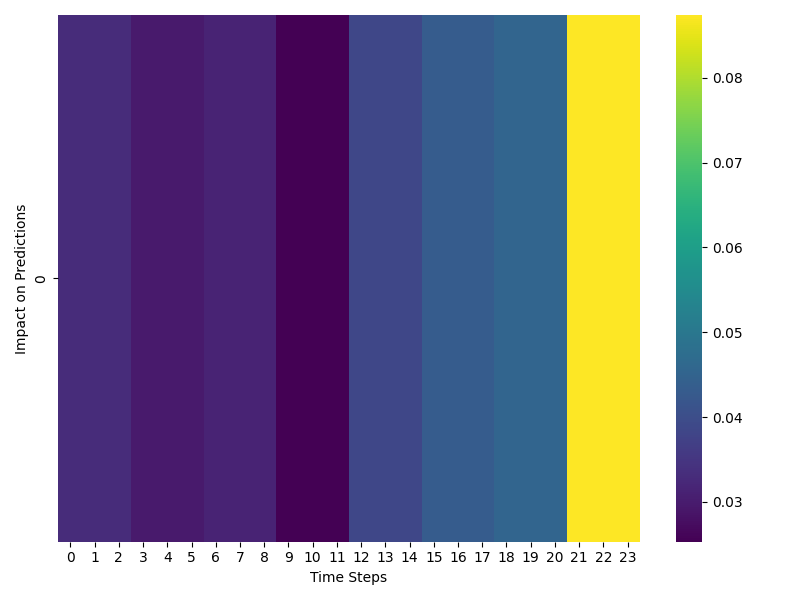}
        \caption{CrossScaleNet ShapTime}
        \label{fig:subfig1}
    \end{subfigure}
    \begin{subfigure}[t]{0.25\textwidth}
        \centering
        \includegraphics[width=\textwidth]{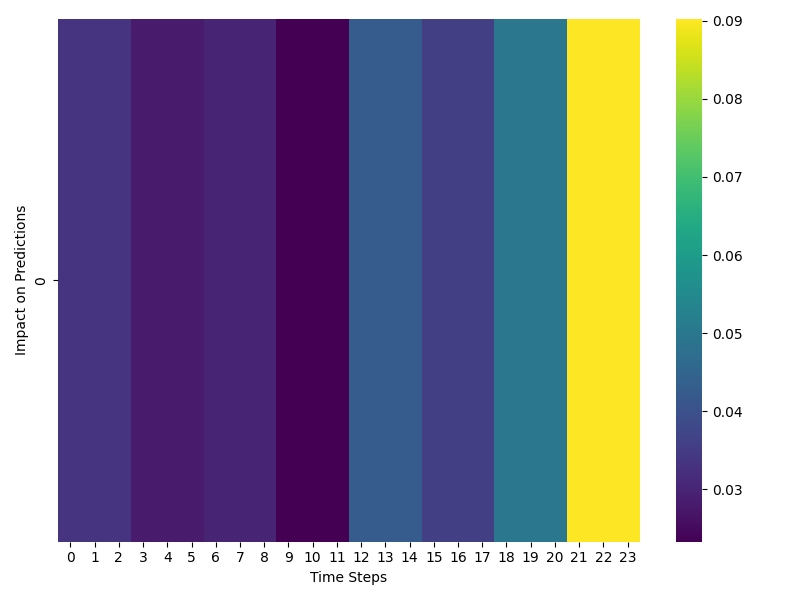}
        \caption{iTransformer ShapTime}
        \label{fig:subfig2}
    \end{subfigure}
    \begin{subfigure}[t]{0.25\textwidth}
        \centering
        \includegraphics[width=\textwidth]{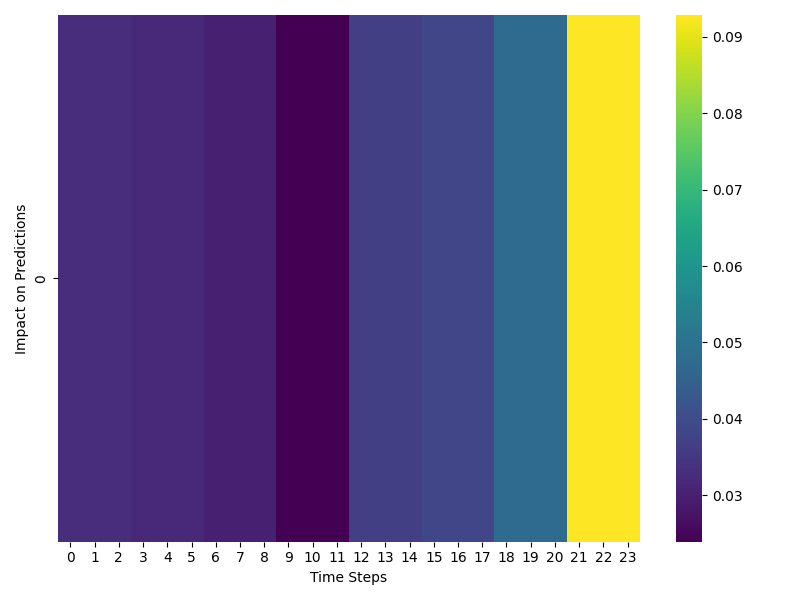}
        \caption{PatchTST ShapTime}
        \label{fig:subfig3}
    \end{subfigure}

    \vspace{1em} 
    \begin{subfigure}[t]{0.25\textwidth}
        \centering
        \includegraphics[width=\textwidth]{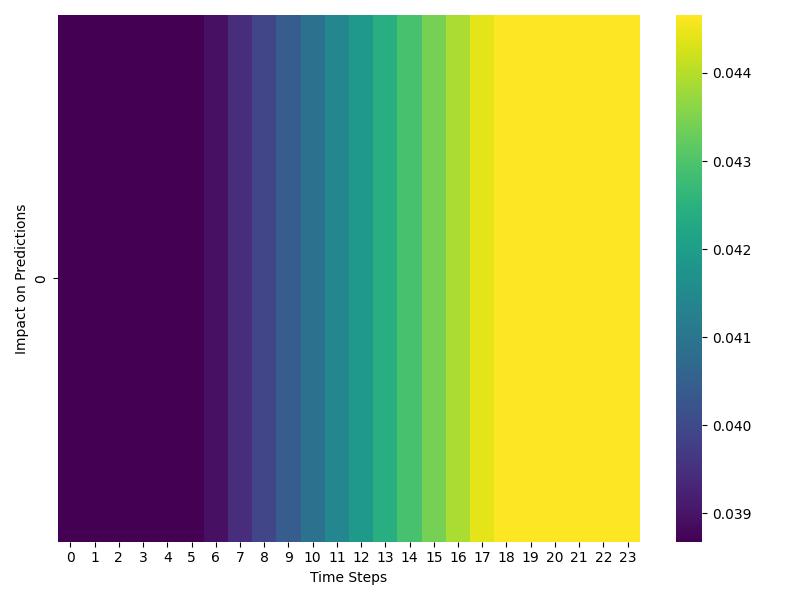}
        \caption{CrossScaleNet attention}
        \label{fig:subfig4}
    \end{subfigure}
    \begin{subfigure}[t]{0.25\textwidth}
        \centering
        \includegraphics[width=\textwidth]{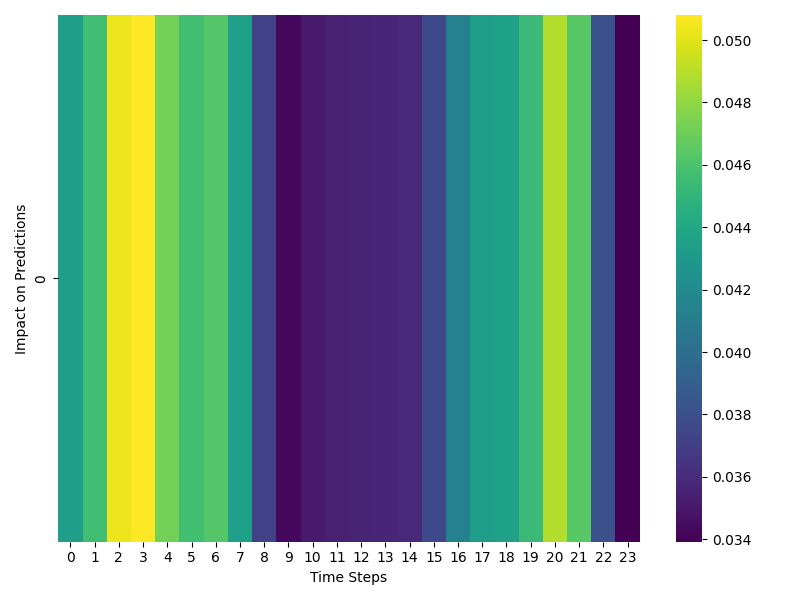}
        \caption{iTransformer attention}
        \label{fig:subfig5}
    \end{subfigure}
    \begin{subfigure}[t]{0.25\textwidth}
        \centering
        \includegraphics[width=\textwidth]{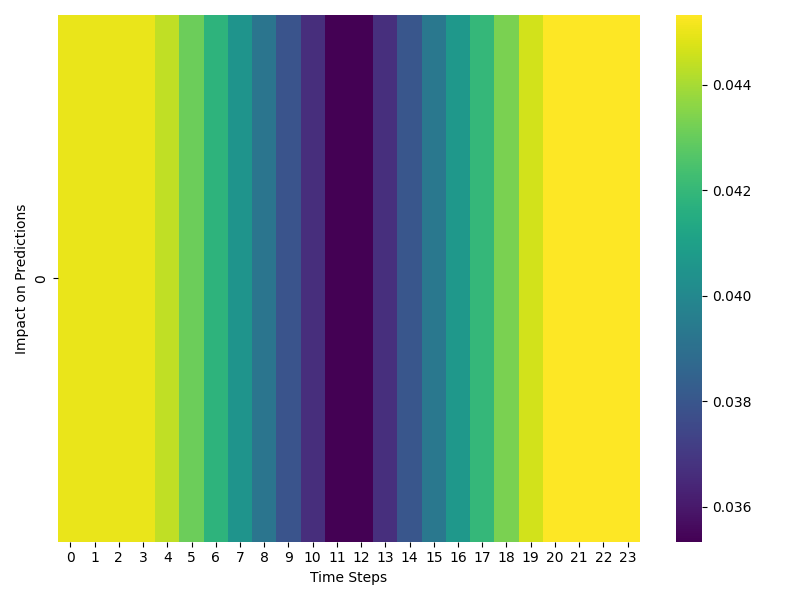}
        \caption{PatchTST attention}
        \label{fig:subfig6}
    \end{subfigure}
\vspace{1.5em}
\caption{ShapTime and attention comparison on PM2.5 dataset.}
\label{fig:PM25}
\end{figure*}

\subsection{Comparative Analysis of Explainability Methods} \label{section:explain}

\begin{figure*}[ht] 
  \centering
    \includegraphics[width=0.9\textwidth]{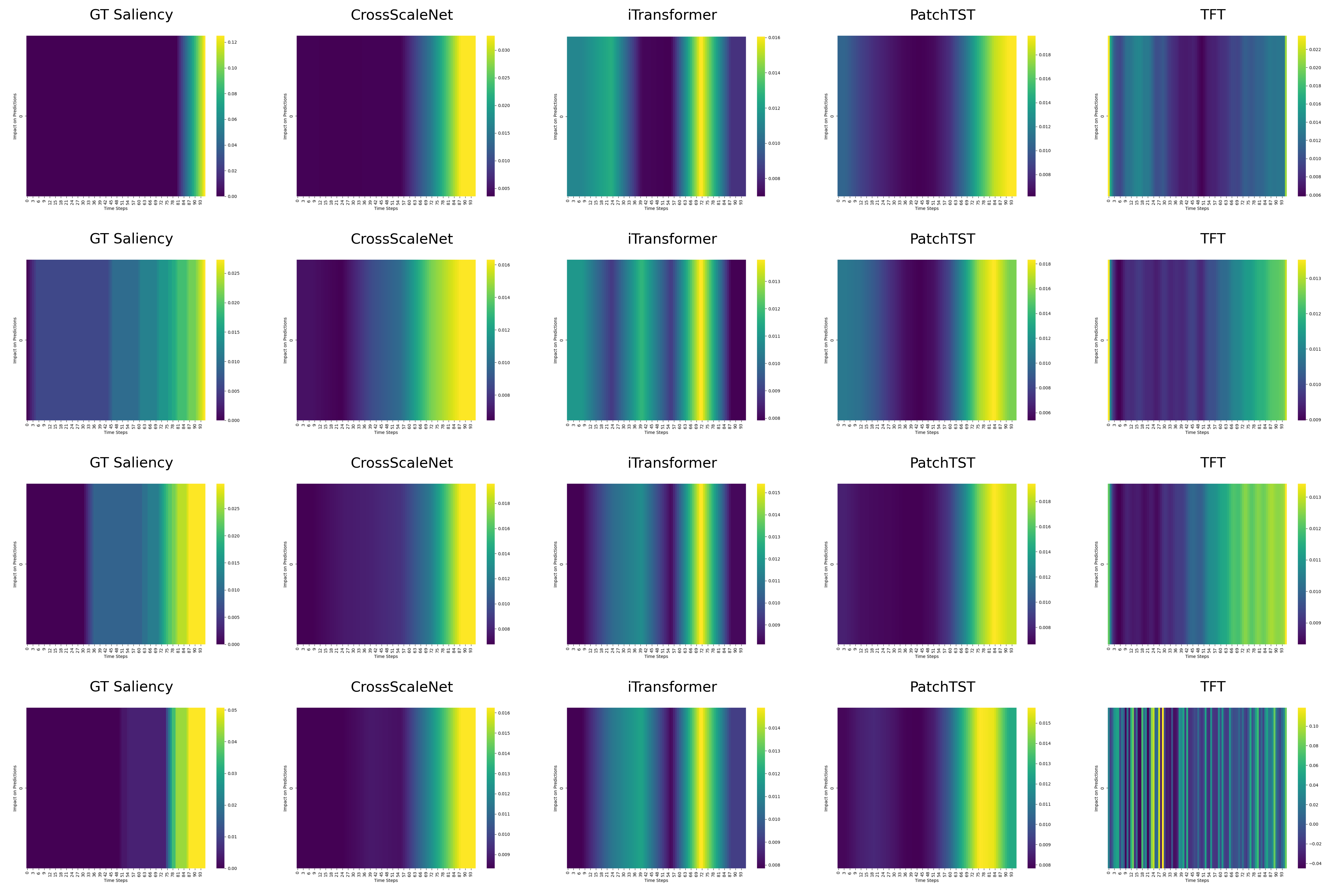}
    \caption{Visualization of ground truth and predicted saliency maps via different architectures for synthetic datasets: SYN1-2-3-4}
    \label{fig:temporal_saliency1}
\end{figure*}

\begin{figure*}[ht] 
  \centering
    \includegraphics[width=0.9\textwidth]{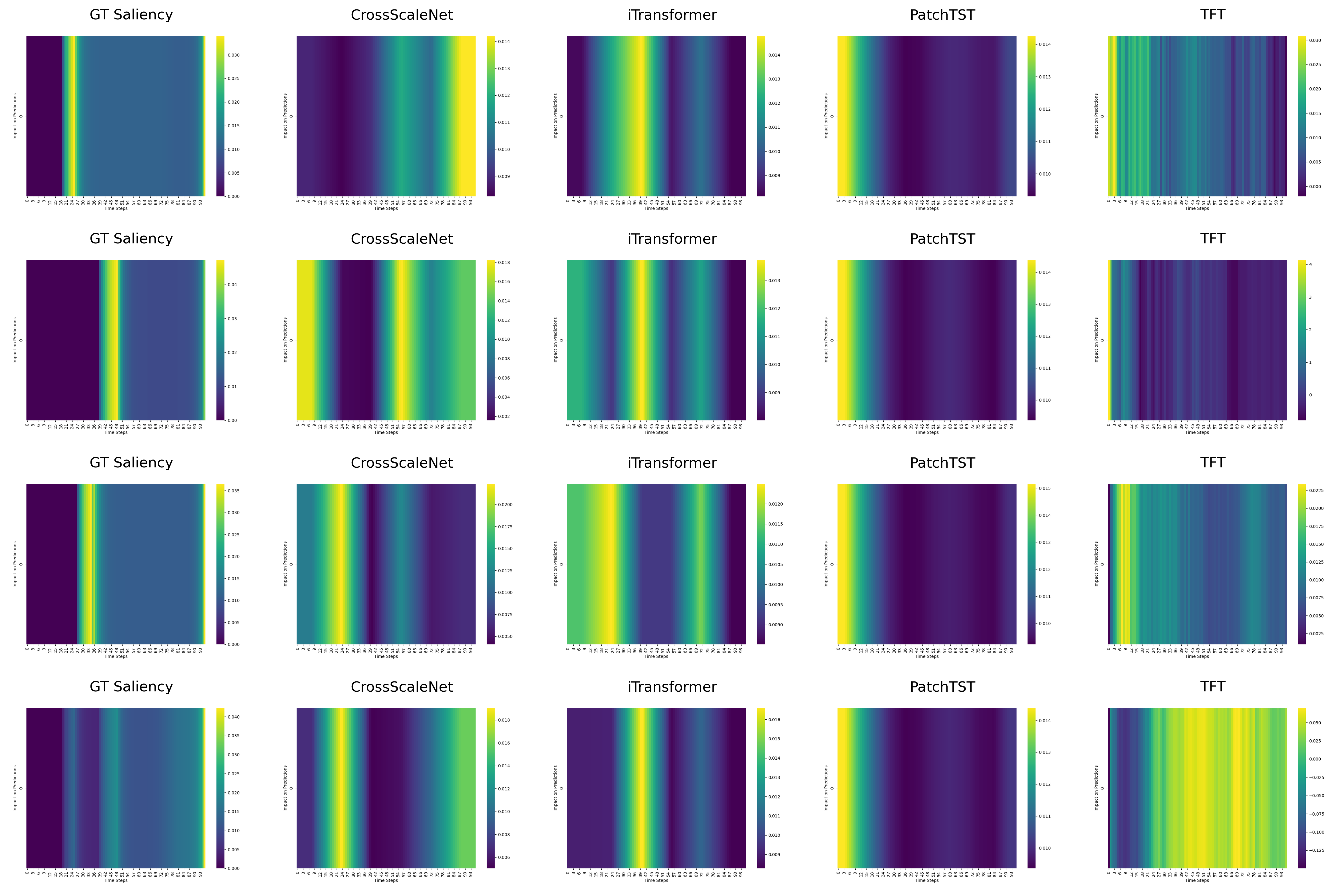}
  \caption{Visualization of ground truth and predicted saliency maps via different architectures for synthetic datasets: SYN5-6-7-8}
  \label{fig:temporal_saliency2}
\end{figure*}

Figure \ref{fig:temporal_saliency1} and \ref{fig:temporal_saliency2} illustrate the temporal saliency maps generated by the proposed CrossScaleNet, iTransformer, PatchTST, and TFT models. PatchTST's attention weights roughly capture temporal saliency, focusing on both recent and distant time points. Since it processes each channel independently, the model uses only the target's own history, without incorporating information from other features. TFT struggles particularly with SYN1 and SYN4, but it may capture temporal saliency relatively better for the other datasets. But its forecasting performance is not good enough as seen in the synthetic dataset and other real-world datasets.iTransformer can not capture temporal saliency on synthetic dataset as seen from the attention map visualization. On the other hand, proposed CrossScaleNet is able to capture temporal saliency in both the last and middle time steps, even if it does not always pinpoint the exact time steps perfectly. We can easily say that CrossScaleNet detects temporal saliency with higher sensitivity than the other models. Our experiments demonstrate that perfectly identifying the exact time step importance is challenging. Nevertheless, our solution offers a reasonable approximation of the temporal saliency with a competitive performance on public real-world benchmarks.

Figure \ref{fig:sample_attentions} compares ShapTime with attention maps for sample prediction on real-world dataset ETTh1. ShapTime consistently provides reliable temporal saliency, but but it is computationally heavy. It is model-agnostic, meaning it does not depend on the architecture or internal mechanisms of the model. Attention maps can be influenced by positional or structural biases inherent to the model, potentially leading to less accurate saliency interpretations. The figure shows that our method effectively captures temporal saliency by producing results closely aligned with ShapTime. We have also utilized \textit{Beijing PM2.5} dataset for explainability comparisons by evaluating ShapTime and attention maps. The results, as shown in Figure \ref{fig:PM25}, demonstrate that only CrossScaleNet produces outputs similar to ShapTime for PM2.5 dataset, highlighting the temporal saliency capacity of our approach. It is also important that each method gives same temporal saliency with ShapTime posthoc XAI method.

We analyze feature importance using Feature Ablation, and Integrated Gradients. Figure \ref{fig:feature_imp_simple} shows feature importance scores from post-hoc explainability techniques. These scores demonstrate the ability of models to detect important features in the synthetic dataset. Our results show that our proposed model outperforms others, especially in detecting important features in the challenging SYN5-8 datasets with simple feature ablation, while iTransformer captures the important features better via Integrated Gradients. Our results demonstrate that the proposed model consistently outperforms others, particularly in identifying important features in the challenging SYN5-8 datasets through simple feature ablation. While our model excels at accurate feature detection, TimeMixer struggles with feature identification on these datasets despite using similar multi-scale processing. Meanwhile, iTransformer effectively captures important features using Integrated Gradients; however, this approach incurs significantly higher computational costs compared to feature ablation.

Temporal saliency is quantitatively assessed using the \textit{sufficiency} and \textit{comprehensiveness} metrics on real-world datasets, with varying percentage thresholds applied to identify salient time steps. Table~\ref{tab-suff_vs_comp} reports the results for these two metrics across three time series models CrossScaleNet (ours), iTransformer, and PatchTST evaluated on two datasets: Electricity and PM2.5. All models extract temporal importance using an \textit{intrinsic} approach via their respective attention mechanisms. 

\subsection{Performance Evaluation on Real-World Datasets} 

Table \ref{tab:pm25_comparison} presents the performance comparison for multivariate input with single-target prediction, where our method demonstrates the best overall performance while also offering explainability. Values highlighted in \textcolor{red}{red} and \textcolor{blue}{blue} indicate the best and second-best results, respectively. The evaluation was conducted for long-term forecasting with a fixed look-back window of 96, as detailed in Table \ref{table-results}. Additionally, Table \ref{table:result_PEMS} compares multivariate short-term forecasting performance across different prediction horizons (12, 24, and 48), using the same look-back window.

\section{Conclusion}

Understanding which parts of a time series contribute most to a model's predictions (temporal salience) is critical to improving transparency, confidence, and decision-making in time series forecasts. However, attention-based models often learn positional or structural biases that do not accurately reflect true temporal importance. In this work, we address this challenge by proposing CrossScaleNet, a novel architecture that integrates cross-scale attention to more effectively capture meaningful temporal saliency. Our results on synthetic datasets with known saliency, along with evaluations on established public benchmarks and real-world forecasting tasks, confirm that CrossScaleNet consistently outperforms state-of-the-art transformer-based models. Importantly, it achieves this while providing intrinsic interpretability, offering actionable insights without compromising predictive accuracy.

\begin{table*}[htb!]
\centering
\scriptsize
\captionsetup{justification=centering}
\caption{Comparison of multivariate long-term forecasting results with prediction horizons (96,192,336,720) and fixed look-back 96. }
\begin{tabular}{|c|c|ll|ll|ll|ll|ll|ll|ll|}
\hline
Models      &        & \multicolumn{2}{c|}{CrossScaleNet} & \multicolumn{2}{c|}{LMSAutoTSF} & \multicolumn{2}{l|}{TimeMixer} & \multicolumn{2}{c|}{iTransformer} & \multicolumn{2}{c|}{PatchTST} & \multicolumn{2}{c|}{ETSformer} & \multicolumn{2}{c|}{TFT} \\ \hline
Database    & Metric & MSE      & MAE    & MSE    & MAE   & MSE   & MAE  & MSE    & MAE    & MSE  & MAE  & MSE   & MAE  & MSE   & MAE  \\ \hline
ETTh1       & 96     & 0.393    & 0.401   & 0.383  & 0.397 & 0.378 & 0.399& 0.386  & 0.404  & 0.386& 0.405 & 0.521 & 0.509 & 0.477 & 0.477 \\
   & 192    & 0.445    & 0.429   & 0.435  & 0.426 & 0.440 & 0.431& 0.443  & 0.434  & 0.447 & 0.440& 0.621 & 0.593 & 0.493 & 0.482 \\
   & 336    & 0.483    & 0.448   & 0.469  & 0.442 & 0.499 & 0.459& 0.492  & 0.463  & 0.489 & 0.473& 0.721 & 0.692 & 0.521 & 0.495 \\
   & 720    & 0.490    & 0.473   & 0.478  & 0.464 & 0.549 & 0.510& 0.508  & 0.491  & 0.506 & 0.496& 0.579 & 0.534 & 0.632 & 0.555 \\ \hline
\textbf{Avg}&        & \textcolor{blue}{0.453}    & \textcolor{blue}{0.438}   & \textcolor{red}{0.441}  & \textcolor{red}{0.432} & 0.466 & 0.449 & 0.457  & 0.448  & 0.457& 0.453 & 0.610 & 0.582 & 0.531 & 0.502 \\ \hline
ETTh2       & 96     & 0.296    & 0.344   & 0.293  & 0.344 & 0.292 & 0.343& 0.294  & 0.346  & 0.318& 0.361 & 0.439 & 0.452 & 0.350 & 0.382 \\
   & 192    & 0.381    & 0.398   & 0.368  & 0.393 & 0.383 & 0.403& 0.379  & 0.398  & 0.389& 0.407& 0.369 & 0.436 & 0.446 & 0.440 \\
   & 336    & 0.413    & 0.427   & 0.411  & 0.427 & 0.439 & 0.437& 0.434  & 0.435  & 0.428& 0.439& 0.498 & 0.451 & 0.459 & 0.456 \\
   & 720    & 0.437    & 0.451   & 0.432  & 0.447 & 0.441 & 0.450& 0.467  & 0.466  & 0.439& 0.454& 0.459 & 0.481 & 0.478 & 0.468 \\ \hline
\textbf{Avg}&        & \textcolor{blue}{0.382}    & \textcolor{blue}{0.405}   & \textcolor{red}{0.376}  & \textcolor{red}{0.403} & 0.389 & 0.408 & 0.394  & 0.411  & 0.394 & 0.415 & 0.441 & 0.455 & 0.433 & 0.437 \\ \hline
ETTm1       & 96     & 0.326    & 0.365   & 0.318  & 0.357 & 0.323 & 0.362 & 0.341  & 0.376  & 0.324& 0.364 & 0.201 & 0.269 & 0.402 & 0.422 \\
   & 192    & 0.378    & 0.394   & 0.357  & 0.378 & 0.368 & 0.386& 0.381  & 0.395  & 0.345& 0.386& 0.293 & 0.342 & 0.453 & 0.449 \\
   & 336    & 0.428    & 0.422   & 0.385  & 0.403 & 0.401 & 0.406& 0.419  & 0.418  & 0.395& 0.408& 0.337 & 0.391 & 0.513 & 0.465 \\
   & 720    & 0.484    & 0.452   & 0.450  & 0.441 & 0.431 & 0.441& 0.486  & 0.455  & 0.397& 0.409& 0.388 & 0.435 & 0.588 & 0.513 \\ \hline
\textbf{Avg}&        & 0.404    & 0.408   & 0.377  & 0.394 & 0.380 & 0.398 & 0.406  & 0.411  & \textcolor{blue}{0.365} & \textcolor{blue}{0.391} & \textcolor{red}{0.304} & \textcolor{red}{0.359} & 0.489 & 0.462 \\ \hline
ETTm2       & 96     & 0.174    & 0.258   & 0.171  & 0.254 & 0.175 & 0.256& 0.184  & 0.266  & 0.186& 0.268& 0.189 & 0.280 & 0.196 & 0.278 \\
   & 192    & 0.240    & 0.300   & 0.236  & 0.298 & 0.238 & 0.299& 0.252  & 0.311  & 0.253& 0.311& 0.253 & 0.319 & 0.258 & 0.315 \\
   & 336    & 0.299    & 0.339   & 0.295  & 0.337 & 0.299 & 0.339& 0.313  & 0.349  & 0.324& 0.358& 0.314 & 0.357 & 0.327 & 0.356 \\
   & 720    & 0.398    & 0.398   & 0.406  & 0.402 & 0.401 & 0.405& 0.409  & 0.405  & 0.406& 0.402& 0.414 & 0.413 & 0.442 & 0.418 \\ \hline
\textbf{Avg}&        & \textcolor{blue}{0.278}    & \textcolor{blue}{0.324}   & \textcolor{red}{0.277}  & \textcolor{red}{0.323} & \textcolor{blue}{0.278} & 0.325 & 0.290  & 0.333  & 0.292 & 0.335 & 0.293 & 0.342 & 0.306 & 0.342 \\ \hline
Weather     & 96     & 0.158    & 0.205   & 0.154  & 0.202 & 0.162 & 0.208& 0.179  & 0.219  & 0.175& 0.218& 0.221 & 0.286 & 0.202 & 0.250 \\
   & 192    & 0.209    & 0.252   & 0.202  & 0.247 & 0.207 & 0.251& 0.225  & 0.258  & 0.222& 0.257& 0.269 & 0.302 & 0.254 & 0.286 \\
   & 336    & 0.271    & 0.295   & 0.259  & 0.288 & 0.263 & 0.292& 0.279  & 0.298  & 0.279& 0.298& 0.271 & 0.334 & 0.300 & 0.319 \\
   & 720    & 0.356    & 0.351   & 0.338  & 0.340 & 0.343 & 0.344& 0.328  & 0.351  & 0.353& 0.346& 0.294 & 0.357 & 0.373 & 0.365 \\ \hline
\textbf{Avg}&        & 0.249    & 0.276   & \textcolor{red}{0.238}  & \textcolor{red}{0.269} & \textcolor{blue}{0.244} & \textcolor{blue}{0.274} & 0.253  & 0.282  & 0.257 & 0.280 & 0.264 & 0.320 & 0.282 & 0.305 \\ \hline
Electricity         & 96     & 0.155    & 0.260   & 0.154  & 0.252 & 0.158 & 0.249& 0.149  & 0.240  & 0.180& 0.273& 0.187 & 0.304 & 0.184 & 0.289 \\
   & 192    & 0.171    & 0.272   & 0.165  & 0.260 & 0.169 & 0.259& 0.164  & 0.255  & 0.194& 0.289& 0.199 & 0.315 & 0.204 & 0.307 \\
   & 336    & 0.192    & 0.290   & 0.177  & 0.274 & 0.189 & 0.281& 0.183  & 0.275  & 0.217& 0.322& 0.212 & 0.329 & 0.214 & 0.317 \\
   & 720    & 0.217    & 0.307   & 0.204  & 0.299 & 0.228 & 0.314& 0.228  & 0.312  & 0.258& 0.352& 0.233 & 0.345 & 0.230 & 0.329 \\ \hline
\textbf{Avg}&        & 0.184   & 0.282   & \textcolor{red}{0.175}  & \textcolor{blue}{0.271} & 0.186 & 0.276 & \textcolor{blue}{0.181}  & \textcolor{red}{0.270}  & 0.212 & 0.309 & 0.208 & 0.323 & 0.208 & 0.311 \\ \hline
Traffic     & 96     & 0.495    & 0.327   & 0.485  & 0.330 & 0.469 & 0.291& 0.417  & 0.287  & 0.526& 0.347& 0.607 & 0.392 & 0.639 & 0.371 \\
   & 192    & 0.495    & 0.327   & 0.483  & 0.322 & 0.480 & 0.299& 0.432  & 0.292  & 0.522& 0.332& 0.621 & 0.399 & 0.671 & 0.382 \\
   & 336    & 0.508    & 0.333   & 0.493  & 0.323 & 0.498 & 0.301& 0.449  & 0.305  & 0.531& 0.339& 0.622 & 0.396 & 0.676 & 0.383 \\
   & 720    & 0.545    & 0.352   & 0.524  & 0.336 & 0.506 & 0.313& 0.481  & 0.318  & 0.552& 0.352& 0.632 & 0.396 & 0.699 & 0.389 \\ \hline
\textbf{Avg}&        & 0.511    & 0.335   & 0.496  & 0.328 & \textcolor{blue}{0.488} & \textcolor{red}{0.301} & \textcolor{red}{0.445}  & \textcolor{red}{0.301}  & 0.533 & 0.343 & 0.621 & 0.396 & 0.671 & 0.381 \\ \hline
Exchange    & 96     & 0.087    & 0.204   & 0.082  & 0.199 & 0.088 & 0.208& 0.098  & 0.226  & 0.097& 0.216& 0.085 & 0.204 & 0.134 & 0.260 \\
   & 192    & 0.180    & 0.301   & 0.174  & 0.296 & 0.184 & 0.307& 0.188  & 0.312  & 0.182& 0.303& 0.182 & 0.303 & 0.242 & 0.352 \\
   & 336    & 0.342    & 0.424   & 0.324  & 0.410 & 0.376 & 0.442& 0.337  & 0.422  & 0.346& 0.425& 0.348 & 0.428 & 0.439 & 0.488 \\
   & 720    & 0.877    & 0.703   & 0.834  & 0.686 & 0.880 & 0.701& 0.885  & 0.714  & 0.950& 0.731& 1.025 & 0.774 & 1.025 & 0.776 \\ \hline
\textbf{Avg}&        & \textcolor{blue}{0.372}    & \textcolor{blue}{0.408}   & \textcolor{red}{0.354}  & \textcolor{red}{0.398} & 0.382 & 0.415 & 0.377  & 0.419  & 0.394 & 0.419 & 0.410 & 0.427 & 0.460 & 0.469 \\ \hline

Solar    & 96  & 0.214  & 0.264 & 0.214  & 0.264 & 0.259 & 0.293 & 0.228  & 0.271 & 0.236 & 0.287 & 0.573 & 0.595 & 0.268 & 0.311 \\
   & 192  & 0.247  & 0.285   & 0.251  & 0.287 & 0.276 & 0.312 & 0.267  & 0.293  & 0.271 & 0.303 & 0.652 & 0.660 & 0.294 & 0.316 \\
   & 336  & 0.261  & 0.298   & 0.270  & 0.299 & 0.296 & 0.326 & 0.286  & 0.310  & 0.285 & 0.318 & 0.753 & 0.707 & 0.314 & 0.325 \\
   & 720  & 0.261  & 0.296   & 0.269  & 0.302 & 0.297 & 0.318 & 0.301  & 0.320  & 0.283 & 0.317 & 0.761 & 0.718 & 0.321 & 0.339 \\ \hline
\textbf{Avg}&  & \textcolor{red}{0.246}  & \textcolor{red}{0.286} & \textcolor{blue}{0.251}  & \textcolor{blue}{0.288} & 0.282 & 0.312 & 0.271  & 0.299  & 0.269 & 0.306 & 0.685 & 0.670 & 0.299 & 0.323 \\ \hline
\textbf{\begin{tabular}[c]{@{}c@{}}Overall \\ Avg\end{tabular}} &        &  0.342  &  \textcolor{blue}{0.351}  & \textcolor{red}{0.332}  & \textcolor{red}{0.345} & 0.344 & \textcolor{blue}{0.351} & \textcolor{blue}{0.341}  & 0.353  &  0.353 & 0.361 & 0.426 & 0.430 & 0.409 & 0.392 \\ \hline
\end{tabular}
\label{table-results}
\end{table*}

\begin{table*}[hb!]
\scriptsize
\centering
\caption{Comparison of multivariate short-term forecasting results with different prediction horizons (12,24,48) and fixed look-back 96.}
\begin{tabular}{|l|c|cc|cc|cc|cc|cc|cc|c|}
\hline
Models      &        & \multicolumn{2}{c|}{CrossScaleNet} & \multicolumn{2}{c|}{LMSAutoTSF} & \multicolumn{2}{c|}{TimeMixer} & \multicolumn{2}{c|}{iTransformer} & \multicolumn{2}{c|}{PatchTST} \\ \hline
Database    & Metric & MSE      & MAE    & MSE    & MAE   & MSE   & MAE  & MSE    & MAE    & MSE  & MAE \\ \hline
\multirow{6}{*}{PEMS3}   & 12 & \textcolor{red}{0.0651} & \textcolor{red}{0.1699} & \textcolor{blue}{0.0653} & \textcolor{blue}{0.1703} & 0.0656 & 0.1720 & 0.0733 & 0.1803 & 0.0874 & 0.2028 \\
 & 24 & \textcolor{blue}{0.0889} & \textcolor{red}{0.1987} & \textcolor{red}{0.0877} & \textcolor{red}{0.1987} & 0.0921 & 0.2031 & 0.1062 & 0.2189 & 0.1344 & 0.2477 \\
 & 48 & \textcolor{red}{0.1395} & \textcolor{red}{0.2479} & \textcolor{blue}{0.1435} & \textcolor{blue}{0.2564} & 0.1468 & 0.2610 & 0.1806 & 0.2911 & 0.2372 & 0.3339 \\
 & \textbf{Avg} & \textcolor{red}{0.0978} & \textcolor{red}{0.2055} & \textcolor{blue}{0.0988} & \textcolor{blue}{0.2085} & 0.1015 & 0.2120 & 0.1200 & 0.2301 & 0.1530 & 0.2615 \\ \hline
\multirow{6}{*}{PEMS4}   & 12 & \textcolor{red}{0.0728} & \textcolor{red}{0.1771} & 0.0741 & \textcolor{blue}{0.1786} & \textcolor{blue}{0.0737} & 0.1795 & 0.0939 & 0.2009 & 0.1062 & 0.2216 \\
 & 24 & \textcolor{red}{0.0869} & \textcolor{red}{0.1960} & \textcolor{blue}{0.0894} & \textcolor{blue}{0.1981} & 0.0934 & 0.2081 & 0.1364 & 0.2459 & 0.1701 & 0.2817 \\
 & 48 & \textcolor{red}{0.1154} & \textcolor{red}{0.2293} & 0.1236 & \textcolor{blue}{0.2391} & \textcolor{blue}{0.1232} & 0.2411 & 0.2291 & 0.3286 & 0.3124 & 0.3899 \\
 & \textbf{Avg} & \textcolor{red}{0.0917} & \textcolor{red}{0.2008} & \textcolor{blue}{0.0957} & \textcolor{blue}{0.2053} & 0.0968 & 0.2096 & 0.1531 & 0.2585 & 0.1962 & 0.2977 \\ \hline
\multirow{6}{*}{PEMS7}   & 12 & 0.0652 & 0.1647 & \textcolor{red}{0.0606} & \textcolor{red}{0.1580} & \textcolor{blue}{0.0611} & \textcolor{blue}{0.1582} & 0.0716 & 0.1744 & 0.0817 & 0.1953 \\
 & 24 & \textcolor{red}{0.0835} & \textcolor{red}{0.1835} & 0.0837 & 0.1864 & \textcolor{blue}{0.0836} & \textcolor{blue}{0.1839} & 0.1092 & 0.2171 & 0.1353 & 0.2473 \\
 & 48 & 0.1428 & 0.2594 & \textcolor{blue}{0.1216} & \textcolor{blue}{0.2269} & \textcolor{red}{0.1199} & \textcolor{red}{0.2219} & 0.1839 & 0.2881 & 0.2568 & 0.3452 \\
 & \textbf{Avg} & 0.0972 & 0.2025 & \textcolor{blue}{0.0886} & \textcolor{blue}{0.1904} & \textcolor{red}{0.0882} & \textcolor{red}{0.1880} & 0.1216 & 0.2265 & 0.1579 & 0.2626 \\ \hline
\multirow{6}{*}{PEMS8}   & 12 & 0.0788 & \textcolor{red}{0.1798} & \textcolor{blue}{0.0775} & \textcolor{blue}{0.1801} & \textcolor{red}{0.0769} & 0.1808 & 0.0980 & 0.2055 & 0.0990 & 0.2148 \\
 & 24 & 0.1123 & \textcolor{blue}{0.2140} & \textcolor{red}{0.1090} & \textcolor{red}{0.2136} & \textcolor{blue}{0.1097} & 0.2165 & 0.1559 & 0.2646 & 0.1554 & 0.2649 \\
 & 48 & \textcolor{blue}{0.1793} & \textcolor{red}{0.2718} & \textcolor{red}{0.1757} & \textcolor{blue}{0.2719} & 0.1949 & 0.2951 & 0.2922 & 0.3749 & 0.2757 & 0.3558 \\
 & \textbf{Avg} & \textcolor{blue}{0.1235} & \textcolor{red}{0.2219} & \textcolor{red}{0.1207} & \textcolor{red}{0.2219} & 0.1272 & 0.2308 & 0.1820 & 0.2817 & 0.1767 & 0.3118 \\ \hline
\multicolumn{2}{l}{\textbf{Overall Avg}} & \textcolor{blue}{0.1025} & \textcolor{blue}{0.2077} & \textcolor{red}{0.1010} & \textcolor{red}{0.2065} & 0.1034 & 0.2101 & 0.1442 & 0.2492 & 0.1710 & 0.2834 \\ \hline
\end{tabular}
\label{table:result_PEMS}
\end{table*}

\subsection{Model Complexity}

Table~\ref{tab:complexity} compares the computational requirements of CrossScaleNet against contemporary baselines. While iTransformer and PatchTST demonstrate superior training efficiency (2.012s and 2.024s per epoch respectively), CrossScaleNet achieves competitive performance (4.663s/epoch) with only 0.03G FLOPs matching LMSAutoTSF's theoretical complexity while introducing multi-scale interpretability. Notably, CrossScaleNet maintains practical inference speeds (8.3ms/batch) comparable to TimeMixer (9.1ms), and significantly faster than TFT's 112ms latency. The memory footprint remains stable across most models (336-344MB), with CrossScaleNet showing no significant overhead from its cross-attention mechanisms compared to the 380MB requirement of TFT.

\begin{table*}[hb]
\centering
\caption{Computational requirements comparison across models. Lower values are better for all metrics.}
\label{tab:complexity}
\begin{tabular}{lcccc}
\hline
Models   & \begin{tabular}[c]{@{}c@{}}FLOPs \\ (G)\end{tabular} & \begin{tabular}[c]{@{}c@{}}Training Time \\ (seconds for 1 epoch)\end{tabular} & \begin{tabular}[c]{@{}c@{}}Inference Latency \\ (ms for one batch)\end{tabular} & \begin{tabular}[c]{@{}c@{}}GPU Memory \\ (MB)\end{tabular} \\ \hline
CrossScaleNet & 0.03  & 4.663  & 8.3 & 336  \\ \hline
LMSAutoTSF    & 0.03 & 4.858  & 6.5& 336  \\ \hline
TimeMixer & 0.13 & 5.341  & 9.1 & 344  \\ \hline
iTransformer & 0.01  & 2.012  & 3.6  & 262  \\ \hline
PatchTST   & 0.05  & 2.024  & 3.6  & 338  \\ \hline
TFT  & 0.47   & 5.827  & 112  & 380  \\ \hline
\end{tabular}
\end{table*}

  \hfill

\section{Future work}
This work has focused on modeling temporal saliency in multivariate time series using a unified attention mechanism. While our approach effectively captures temporal saliency and achieves strong predictive performance, an important aspect that could further enhance the modeling process is the differentiated influence of target and non-target features over time. In many forecasting tasks, recent target values may contribute more strongly to predictions than auxiliary features or long-range history. Exploring architectures that explicitly separate the processing of target and non-target features within the attention mechanism could provide even more fine-grained saliency estimation and potentially improve performance in scenarios where feature relevance varies significantly over time. This direction offers a promising opportunity to refine both the interpretability and accuracy of time series models across complex, real-world datasets.


\section{Data availability}
All the dataset are available online. The code to reproduce this work
is available at https://github.com/mribrahim/LMS-TSF.

\FloatBarrier  
\bibliographystyle{unsrt}

\bibliography{cas-refs}

\end{document}